\documentclass[review]{elsarticle}

\usepackage{hyperref}

\journal{Pattern Recognition}

\usepackage{amsmath,amsfonts,amssymb}
\usepackage[inline, shortlabels]{enumitem}

\usepackage{siunitx}
\usepackage{soul}
\usepackage{microtype}
\usepackage[super]{nth}
\usepackage[table]{xcolor} 
\usepackage{booktabs, multirow}%
\usepackage{url}
\usepackage{xspace}
\usepackage{subcaption}
\usepackage[ruled,vlined]{algorithm2e}
\usepackage{latexsym}
\usepackage{cleveref}
\crefname{section}{Sec.}{Sections}
\crefname{figure}{Fig.}{Figs.}
\crefname{table}{Tab.}{Tabs.}
\crefname{equation}{Eq.}{Eqs.}
\crefname{appsec}{Appendix}{Appendices}

\usepackage{parskip}
\usepackage{url}
\usepackage{graphicx}
\usepackage{cjhebrew}

\makeatletter
\DeclareRobustCommand\onedot{\futurelet\@let@token\@onedot} 
\def\@onedot{\ifx\@let@token.\else.\null\fi\xspace}
\makeatother

\newcommand{\eg}{e.\,g.,\xspace}

\newcommand{\cf}{cf\onedot}
\newcommand{\ie}{i.\,e.,\xspace}
\newcommand{\wrt}{w.\,r.\,t\onedot}

\newcommand{\etal}{\emph{et al.}\xspace}
\newcommand{\vs}{vs\onedot}

\newcommand{\greenten}{\cellcolor{green!10}}
\newcommand{\greenthirty}{\cellcolor{green!30}}
\newcommand{\redten}{\cellcolor{red!10}}

\newcommand{\icc}{ICC$++$\xspace}

\begin{document}

\begin{frontmatter}

\title{ICC++: Explainable Image Retrieval for Art Historical Corpora using Image Composition Canvas}

\author{Prathmesh Madhu\corref{cor1}\fnref{fn1}}
\ead{prathmesh.madhu@fau.de}
\cortext[cor1]{Corresponding author}
\fntext[fn1]{Pattern Recognition Lab, FAU}

\author{Tilman Marquart\fnref{fn1}}
\ead{tilman.marquart@fau.de}

\author{Ronak Kosti\fnref{fn1}}
\ead{ronak.kosti@fau.de}

\author{Dirk Suckow\fnref{fn3}}
\ead{dirk.suckow@fau.de}
\fntext[fn3]{%
  Institute of Art History, FAU
}

\author{Peter Bell\fnref{fn4}}
\ead{peter.bell@uni-marburg.de}
\fntext[fn4]{%
  German Studies and Art Studies, Philipps University of Marburg
}

\author{Andreas Maier\fnref{fn1}}
\ead{andreas.maier@fau.de}

\author{Vincent Christlein\fnref{fn1}}
\ead{vincent.christlein@fau.de}


\begin{abstract}
    Image compositions are helpful in the study of image structures and assist in discovering semantics of the underlying scene portrayed across art forms and styles. With the digitization of artworks in recent years, thousands of images of a particular scene or narrative could potentially be linked together. However, manually linking this data with consistent objectiveness can be a highly challenging and time-consuming task. In this work, we present a novel approach called Image Composition Canvas (\icc) to compare and retrieve images having similar compositional elements. \icc is an improvement over ICC specializing in generating low and high-level features (compositional elements) motivated by Max Imdahl's work. To this end, we present a rigorous quantitative and qualitative comparison of our approach with traditional and state-of-the-art (SOTA) methods showing that our proposed method outperforms all of them. In combination with deep features, our method outperforms the best deep learning-based method, opening the research direction for explainable machine learning for digital humanities. We will release the code and the data post publication.
\end{abstract}


\end{frontmatter}


\section{Introduction}\label{sec:intro}
	Art historians understand images by studying the underlying structures and discovering  the semantics of scenes in historical paintings. 
	One way of understanding is by analyzing the underlying image composition. 
	Prominent art historians in the \nth{20} century have defined various compositional elements~\cite{hetzerGiottoSeineStellung1941,imdahlGiottoArenafreskenIkonographie1996} for image analysis and understanding. 
	Max Imdahl underlined the aesthetic and semantic importance of the structural composition of an image. 
	The strength of this method lies in the abstraction of essential elements from paintings which can then be used to compare them for same or different artists using the underlying common elements~\cite{imdahlGiottoArenafreskenIkonographie1996}. 
	On close observation, it is apparent that these image composition elements rely on the abstraction of the depicted body postures of the portrayed characters. 
	
\begin{figure*}[!t]
		\centering
		\begin{subfigure}{0.28\linewidth}
			\includegraphics[height=3cm,width=2cm]{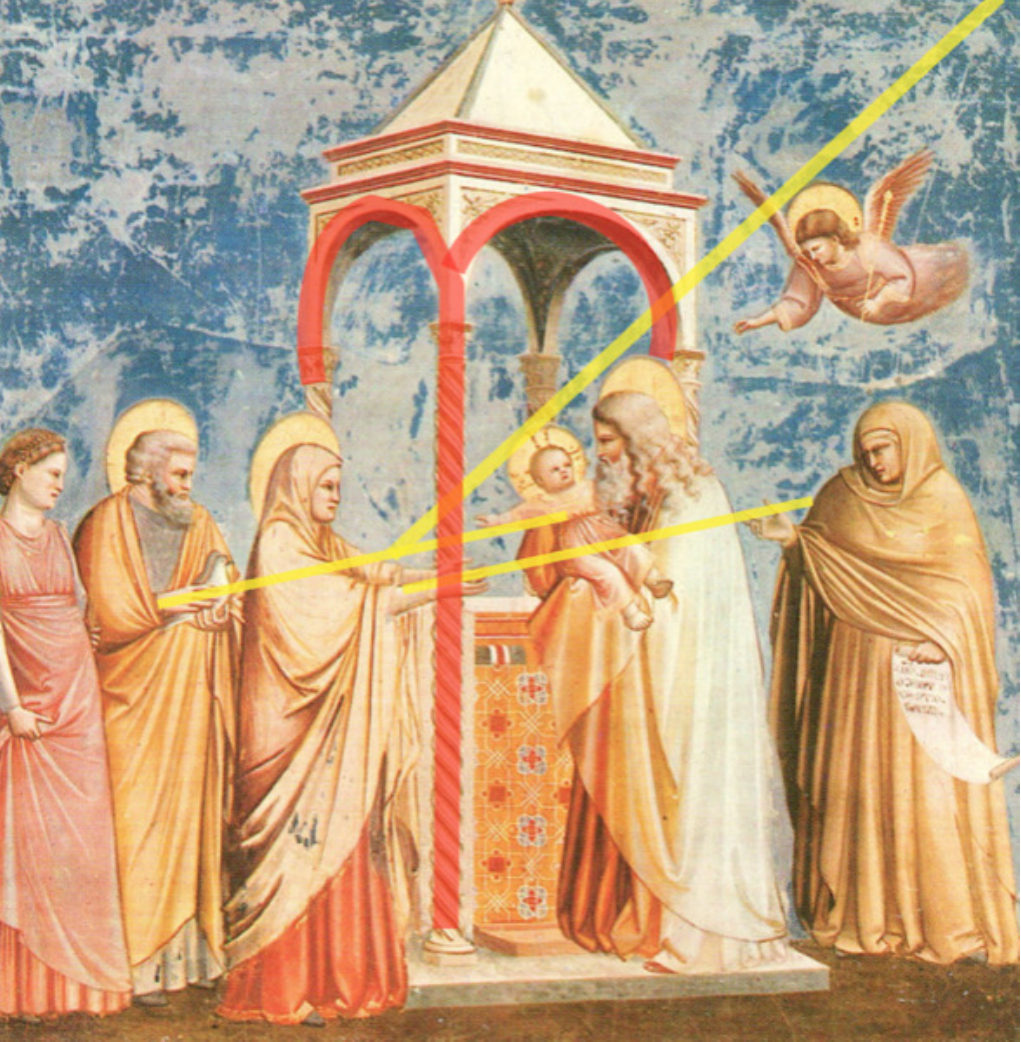}
			\caption{}
			\label{fig:artcompodiagrams_a}
		\end{subfigure}
		\begin{subfigure}{0.28\linewidth}
			\includegraphics[height=3cm]{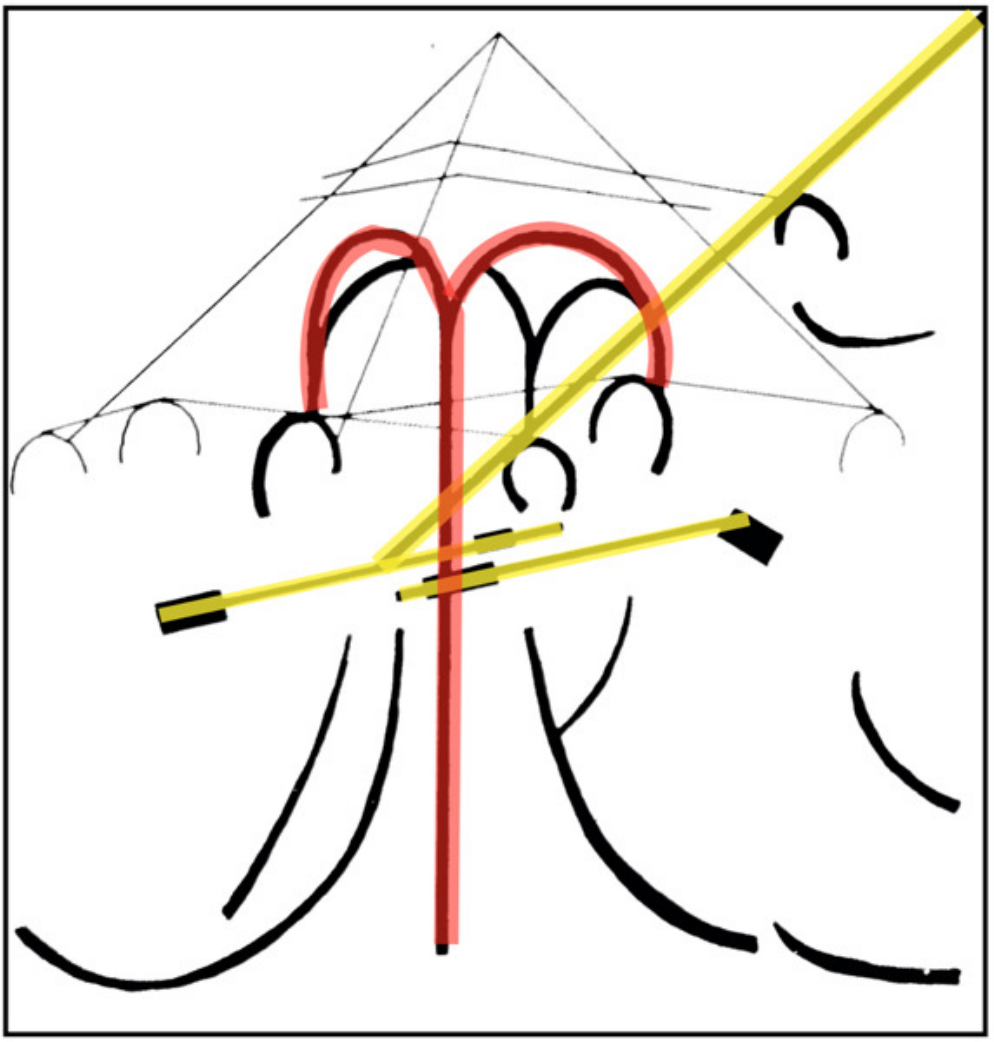}    
			\caption{}
			\label{fig:artcompodiagrams_b}
		\end{subfigure}
		\begin{subfigure}{0.40\linewidth}
			\includegraphics[height=3cm]{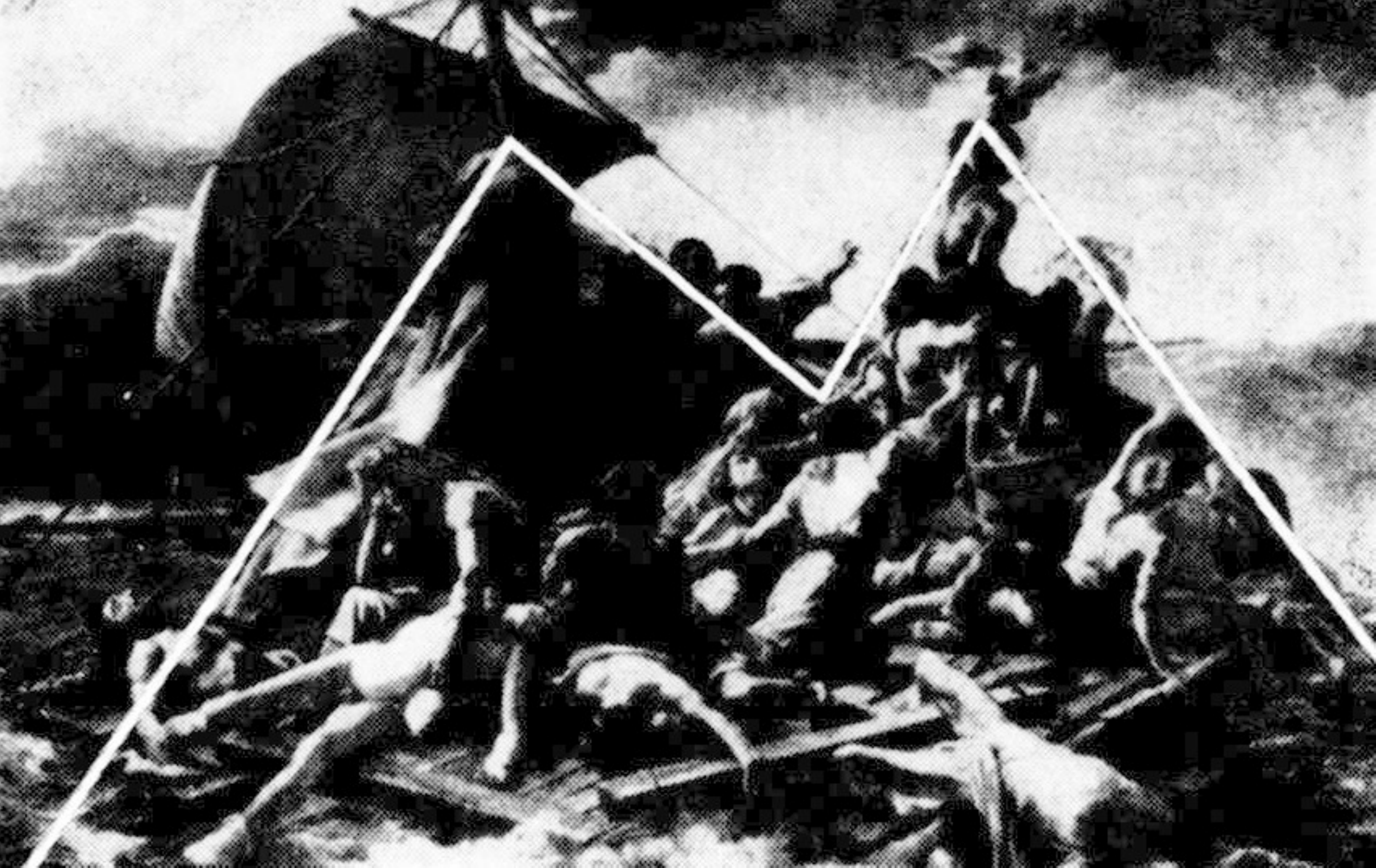}    
			\caption{}
			\label{fig:artcompodiagrams_c}
		\end{subfigure}
		\caption{Understanding compositions in paintings;
			\subref{fig:artcompodiagrams_a} Presentation of Jesus in the Temple by Giotto, 1305, Arena Chapel, Padua; \subref{fig:artcompodiagrams_b} Composition Diagram for \subref{fig:artcompodiagrams_a} by Max Imdahl~\cite{imdahlGiottoArenafreskenIkonographie1996}, 1980; \subref{fig:artcompodiagrams_c}  M-shaped character cluster of the painting \textit{The Raft of the Medusa} by Hans Körner~\cite{kohleHansKornerAuf1990}, 1988; \subref{fig:artcompodiagrams_a} \& \subref{fig:artcompodiagrams_b} View directions, movement \& action patterns in yellow, shapes of specific color areas in red; the two parallel yellow action patterns link the relationship between four central protagonists in the image based on their body orientations.}
		\label{fig:artcompodiagrams}
	\end{figure*}

	\emph{\textbf{Composition in paintings}}: Over the years, many art historians have presented unique approaches for creating image composition diagrams that relate the latent structures and relations between the characters in paintings. One of the very first structured methods to create composition diagrams were introduced by Hetzer~\cite{hetzerGiottoSeineStellung1941} in 1941. Hetzer speaks of a `\textit{mathematical structure}' of painting (by Giotto) and its `\textit{linear relations}'. This shows, on the one hand, possible proximity to computational procedures and, on the other hand, also efforts of art history to be perceived as a science. Dagobert Frey sees a whole development of `\textit{Geometrisierung}' (geometrization) in art history~\cite{frey1952giotto}. Max Imdahl built upon Hetzer's ideas in his main work titled \textit{Ikonik} in 1996~\cite{imdahlGiottoArenafreskenIkonographie1996}, where he studied the composition of seminal works by Giotto di Bondone -- an Italian artist of the \nth{13} century known for his paintings and frescoes, which depicted characters via semantics of their arrangements around one another. In general, these composition diagrams can become very complex as shown in \cref{fig:artcompodiagrams_b} and \cref{fig:artcompodiagrams_c}, many of which are based on the posture abstraction of the visible characters in the image~\cite{imdahlGiottoArenafreskenIkonographie1996,kohleHansKornerAuf1990}. 
	
	Very few works studied the image composition diagrams~\cite{imdahlGiottoArenafreskenIkonographie1996,kohleHansKornerAuf1990}. All of the approaches follow a bottom-up approach, which can briefly be described in two steps: \textit{First}, a distinctive low-level interpretation of the objects and characters in the artwork; and \textit{second}, an abstract high-level interpretation built upon this low-level interpretation. The former step is an easy task for humans as our brain processes this subconsciously when we look at images or paintings~\cite{kapadiaSpatialDistributionContextual2000,longMidlevelVisualFeatures2018}, while in the latter, we look for relations between characters as shown in \cref{fig:artcompodiagrams_a} and \cref{fig:artcompodiagrams_b}. They can be prominent action patterns as seen in \cref{fig:artcompodiagrams_a} and \cref{fig:artcompodiagrams_b} or the M-shaped character clusters in \cref{fig:artcompodiagrams_c}. Finding these relations automatically within a dataset is a highly challenging and time-consuming task, which requires extensive domain knowledge training, which is studied by experts in art history. Bell and Impett~\cite{bellchoreography} also analyze postures of characters to understand the impact of poses on the semantic understanding of an iconography, like the annunciation of the lord. Besides this, various art historians may present subjective composition diagrams caused due to the abstract high-level interpretations. 
	
	The necessity to generate compositions is to retrieve compositionally similar-looking artworks. The advancement in the digitization of artworks (REPLICA project~\cite{seguin2018new,tallon_2017}) has made it possible to link artworks rapidly via computational resources. Finding visual patterns, and understanding and linking them across large databases can be considered of prime importance for art historical research. The importance of content-based image retrieval systems for art historians was highlighted by Seguin \etal~\cite{seguinVisualLinkRetrieval2016}. However, due to the varied understanding of similarities (color, shape, context), this is more challenging than it might seem. Since linking paintings is an N-to-N matching problem, art historians require years of experience and research to develop the skills necessary for finding such connections. Ideally, a content-based retrieval system makes finding connections between artworks better, easier and faster.
	
	In ICC~\cite{madhu2020understanding}, we presented a novel approach for generating composition diagrams. Our method is based on abstraction and grouping of low-level features to generate an abstract higher-level interpretation of any painting using existing pre-trained neural networks. 
	
	In this work, we improve ICC~\cite{madhu2020understanding} and present \icc. Our contributions include:
	\begin{enumerate}
		\item \icc with improvements in each feature of ICC (\cref{subsec:iccdrawbacks}) to generate composition elements for building an explainable content-based image retrieval system based on compositional similarity.
		\item Introducing a small dataset called Web Gallery of Art 500 (or \textit{WGA500}) to evaluate our retrieval method and make it publicly available to encourage further research in this domain.  
		\item \icc for compositional similarity beats the SOTA method LATP~\cite{jenicekLinkingArtHuman2019} for image retrieval.
		\item Rigorous qualitative and quantitative evaluation of our methods against SOTA deep-learning-based retrieval methods.
	\end{enumerate}

    This work is structured as follows: in \cref{sec:relwork}, we review recent computer vision approaches for representation learning and retrieval and briefly mention existing approaches that use compositions for retrieval. \cref{sec:method} introduces the proposed method where we develop a retrieval framework for art historical images using compositional elements. The datasets, evaluation protocol and experimental design are described in \cref{sec:expsetup}. Qualitative and Quantitative results are presented and discussed in \cref{sec:results}. The paper is concluded in \cref{sec:conclusion}.
	
	\section{Related Work}\label{sec:relwork}
	Computer vision research has produced many effective methods for low-level interpretation, which can then be used to devise algorithms further and build higher-level interpretations. 
	Methods based on deep learning using convolutional neural networks~\cite{simonyanVeryDeepConvolutional2015} are predominant and show promising results. 
    These methods enable the detection of low-level features, such as objects, character, or gazes in images.
	Most of these methods claim to have high domain adaptability and thus are candidates to be adapted for their use in artworks. 
	However, recent works~\cite{madhu2019recognizing,caiCrossDepictionProblemComputer2015} have shown that a huge domain gap exists which impacts the performance, and hence we need sophisticated domain adaptation. 
	
	In previous works~\cite{hetzerGiottoSeineStellung1941,imdahlGiottoArenafreskenIkonographie1996}, the body postures of the characters in the paintings were mainly used in describing the compositions of the paintings.
	Gaze direction or body orientation of the characters, \ie mid-level features, are also considered for the generation of compositions. 
	Existing vision methods heavily rely on the visibility of eyes in the image or a video stream to estimate the gaze~\cite{parkFewShotAdaptiveGaze2019,kellnhoferGaze360PhysicallyUnconstrained2019} which severely restricts their direct application or adaption to the art domain. 
	For example, Recasens \etal~\cite{recasens2016they} proposed an end-to-end approach for gaze estimation that requires head and eye position in addition to the input image to predict where the person is looking. 
	This method~\cite{recasens2016they} can be applied to 2-D real-world images, but also require annotations in the testing phase, which presents an additional challenge in adapting these methods to the art domain. 
	Following the low and mid-level features, we~\cite{madhu2020understanding} abstracted the characters using existing pre-trained pose estimation methods to form a low-level interpretation, which in turn are used to automatically generate high-level features called compositional elements.

	Several modern image retrieval systems have been applied to artworks~\cite{seguinVisualLinkRetrieval2016,seguin2017tracking}. 
	The most critical aspect of modern retrieval systems is the concept of image similarity for linking images. 
	The existing content-based retrieval systems like VGG GeM~\cite{radenovicFinetuningCNNImage2018} or Deep Image Retrieval~\cite{gordoDeepImageRetrieval2016} focus on the similarity of images using deep visual features. 
	Seguin~\cite{seguinReplicaProjectBuilding2018} developed a visual search engine for finding replicas in paintings for art history. 
	Deep visual features are computed using triplet loss~\cite{dong2018triplet} to find duplicate or replica images by comparing them in their descriptor space. 
	The use of such deep learning-based features makes an interpretation difficult. 
	In contrast, we present a more interpretable method.
	Related to our task of image retrieval is the work of~\cite{shen2019discovering}, where the authors try to find similar patterns across images rather than total image similarity. 
	Their approach relies on candidate proposals and correspondences based on patch matching and then uses the matched patches to improve the features using metric learning loss. 
    Unfortunately, the approach is a rather slow process, which limits the size of the dataset as mentioned by the authors.  
	
	Due to the above-mentioned challenges, little work has been done in generating the higher-level interpretation or creating and understanding scenes in paintings using image compositions. 
	Jenicek \etal~\cite{jenicekLinkingArtHuman2019} proposed a composition transfer method to find similar pose compositions, which is a superior replacement for content-based image retrieval for a manually annotated collection of specialized artworks. 
	They introduced a pose matching method, called \emph{Linking Art through Human Poses} (LATP), to link artworks based on the pose similarity of the characters. Their work is closely related to our work, however their work differs in two major aspects: 
	(1)~They rely only on the pre-trained SOTA pose estimation methods for generating pose keypoints, whereas we use additional features (\cref{sec:method}); (2) their method is useful for finding similar compositions to discover copies and replications in different media of the same artwork, whereas our method can be applied across different artworks to robustly find influences across mediums, periods, art schools, and artists.

	\begin{figure*}[!t]
		\centering
		\begin{subfigure}{0.24\linewidth}
			\includegraphics[height=3cm]{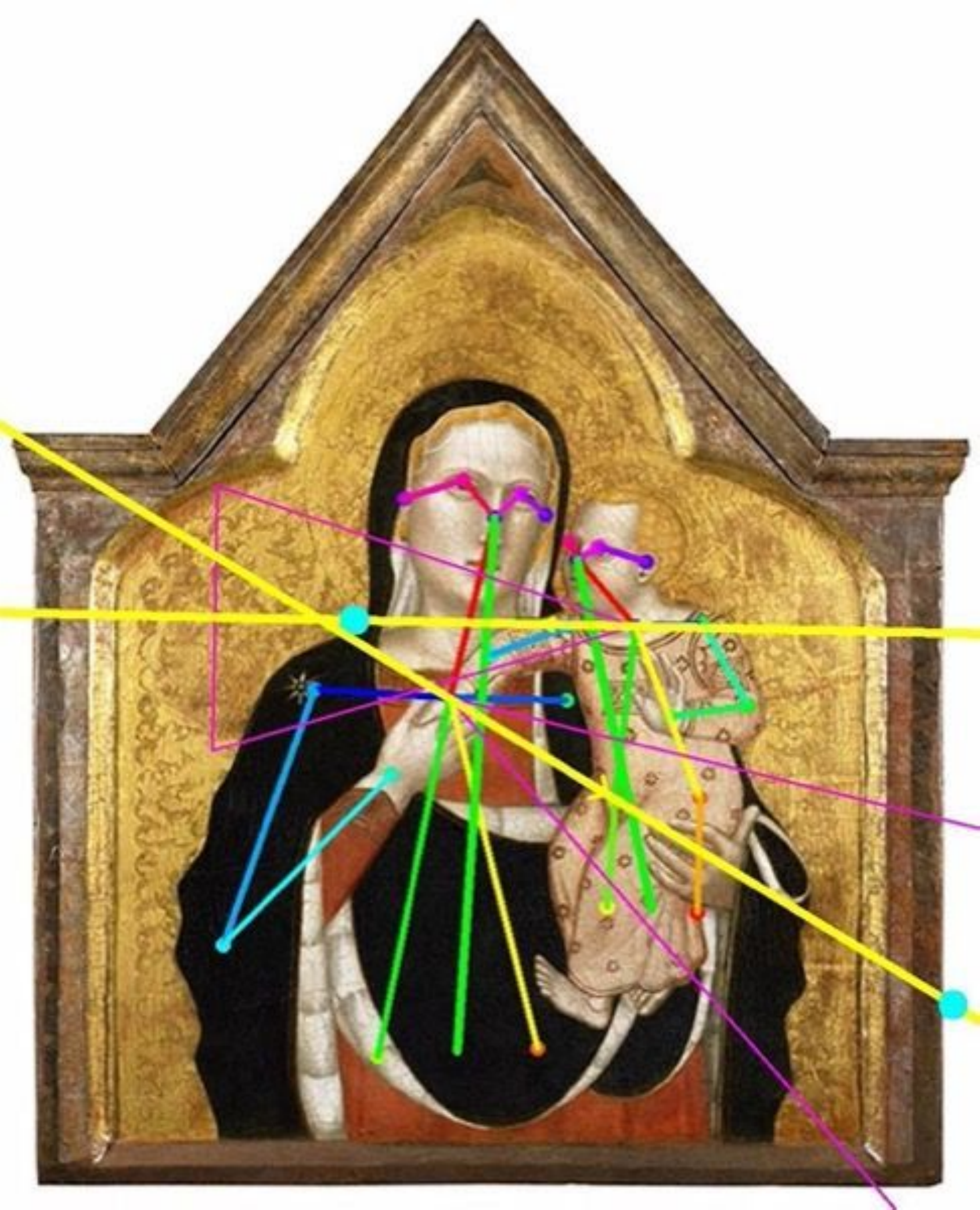}
			\caption{}
			\label{fig:coneissues_a}
		\end{subfigure}
		\begin{subfigure}{0.24\linewidth}
			\includegraphics[height=3cm]{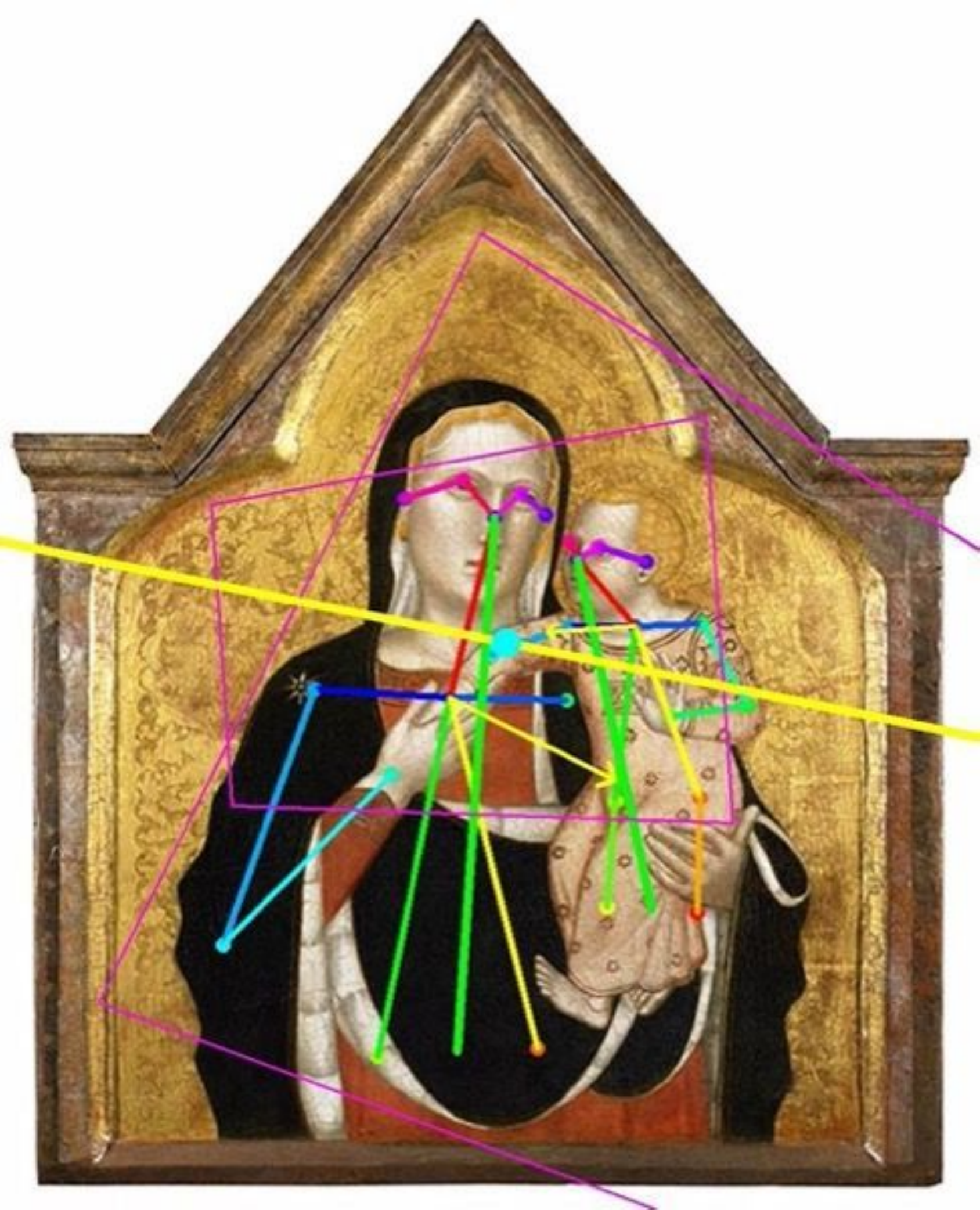}    
			\caption{}
			\label{fig:coneissues_b}
		\end{subfigure}
		\begin{subfigure}{0.24\linewidth}
			\includegraphics[height=3cm]{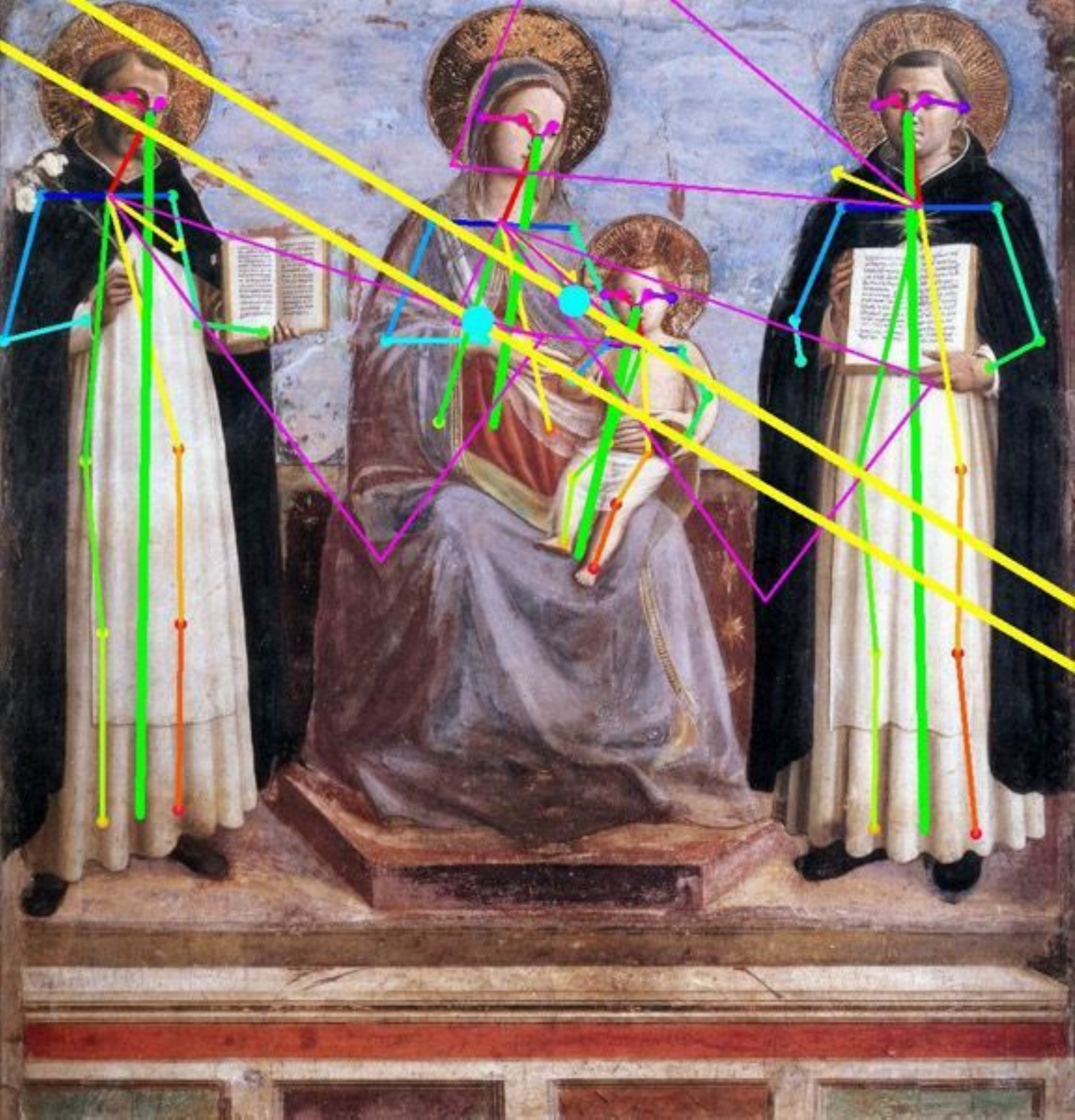}    
			\caption{}
			\label{fig:coneissues_c}
		\end{subfigure}
		\begin{subfigure}{0.24\linewidth}
			\includegraphics[height=3cm]{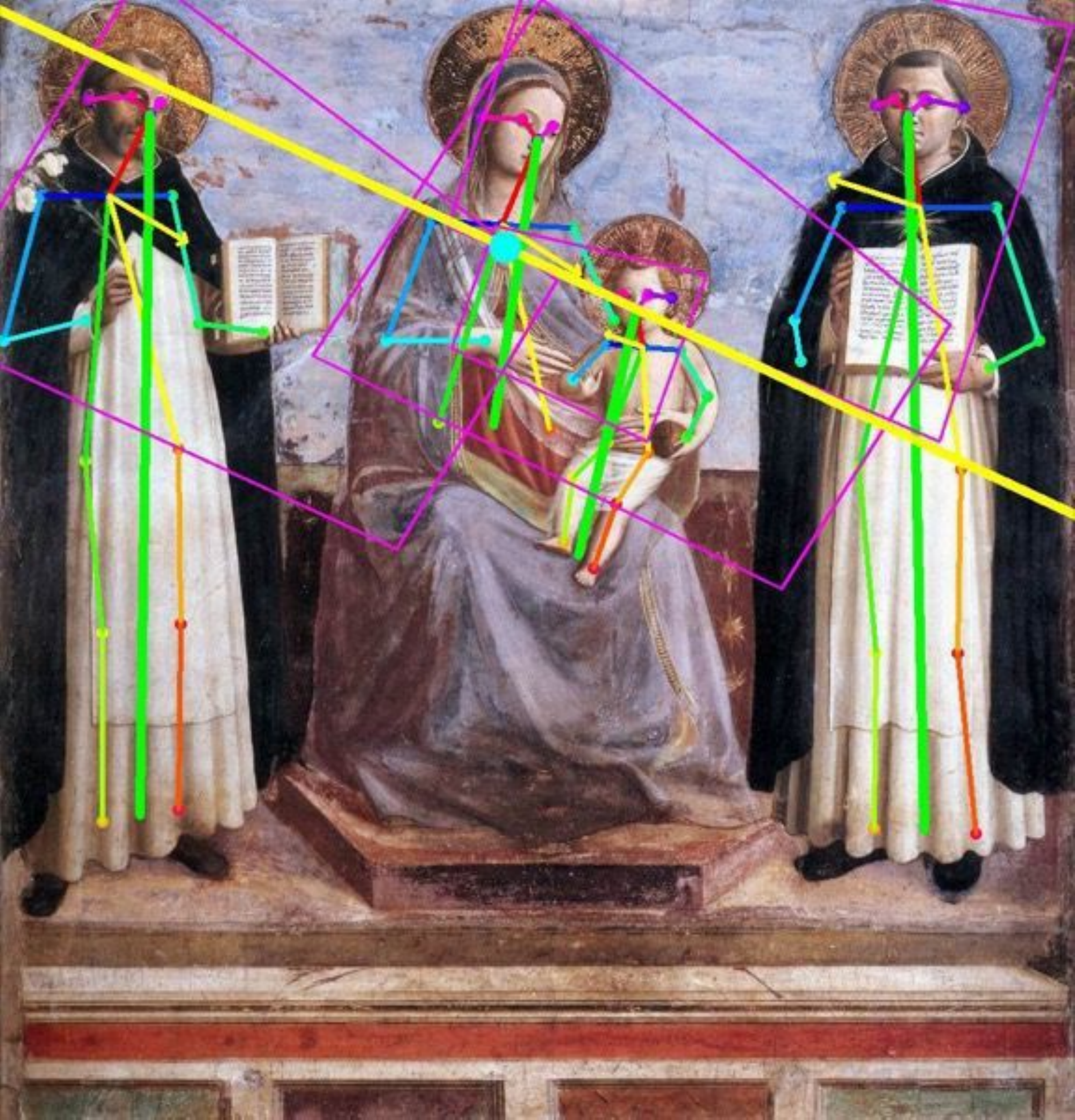}    
			\caption{}
			\label{fig:coneissues_d}
		\end{subfigure}
		\caption{A painting from the \textit{virgin and child} iconography; poselines in green, bisection cone in magenta, global action lines in yellow and global action centers in cyan; 
		\subref{fig:coneissues_a},\subref{fig:coneissues_c} no cone-based scaling, hence multiple incorrect action centers detected; 
		\subref{fig:coneissues_b},\subref{fig:coneissues_d} with cone-based scaling resulting in improved global action centers}
		\label{fig:coneissues}
	\end{figure*}

\section{Improved Image Composition Canvas (ICC++)}\label{sec:method}
	In this section, we first briefly introduce the Image Composition Canvas (ICC) and its drawbacks (\cref{subsec:iccdrawbacks}), then we highlight our improvements in \icc (\cref{subsec:ICC++}), and finally, present an image retrieval system (\cref{subsec:ICC++ret}) based on \icc. 
	
	\subsection{ICC and its drawbacks}\label{subsec:iccdrawbacks}
	Image Composition Canvas (ICC) was first introduced by Madhu \etal~\cite{madhu2020understanding}, which uses pre-trained deep neural networks for low-level interpretation of the image content, which are subsequently used to generate higher-level interpretations. 
	ICC allows visual abstraction of the inherent structural semantics of the scene, which allows for more rigorous scene analysis. In addition, these abstractions can be quantified and used as features to compare different images. Short definitions of these abstractions in the form of compositional elements are given in Supplementary (Sec. 7.1). 
	Despite the method being effective and validated by a user study, the ICC pipeline has a few drawbacks.
	
	\begin{enumerate}
		\item \label{subsec:limit-triangle-abstraction} \textbf{Triangle Abstraction:} While using the pre-trained pose estimators, oftentimes the keypoints for the lower body part are not detected because either the method fails or the lower part is not visible. 
	    For example, this problem occurs for the \textit{virgin and child} iconography as seen in \cref{fig:coneissues_a}, where the character of Mary is only half visible in the paintings. 
		
		\item \label{subsec:limit-cone-length} \textbf{Cone issue:} While calculating the action regions (\cf the cyan dots in \cref{fig:coneissues}), the character sizes were not considered.
			This resulted in a shift of the global action regions for groups of small characters in an image. 
			To compute the action region, a body direction is first computed (from the body pose) and then a cone is drawn in this direction which starts with size zero from the body-origin. 
			Hence, when the cones from multiple characters intersect, the body direction can slightly miss the other cones resulting in multiple intersecting regions (and therefore more action centers) than actually present in the image. This can be observed in \cref{fig:coneissues_a} and \cref{fig:coneissues_c}.
		
		\item \label{subsec:limits-bisection-bug} \textbf{Bisection angle issue:} If any character in the image is looking up (down), ICC fails at detecting the correct direction for the bisection angle (and therefore the body direction) resulting in many false positives for the global action lines and global action centers.
			A simulation of incorrect bisection angles is shown in Supplementary (Fig. 16) in magenta. 
	\end{enumerate}

	\begin{figure*}[!t]
		\centering
		\begin{subfigure}{0.2\textwidth}
			\includegraphics[width=\textwidth]{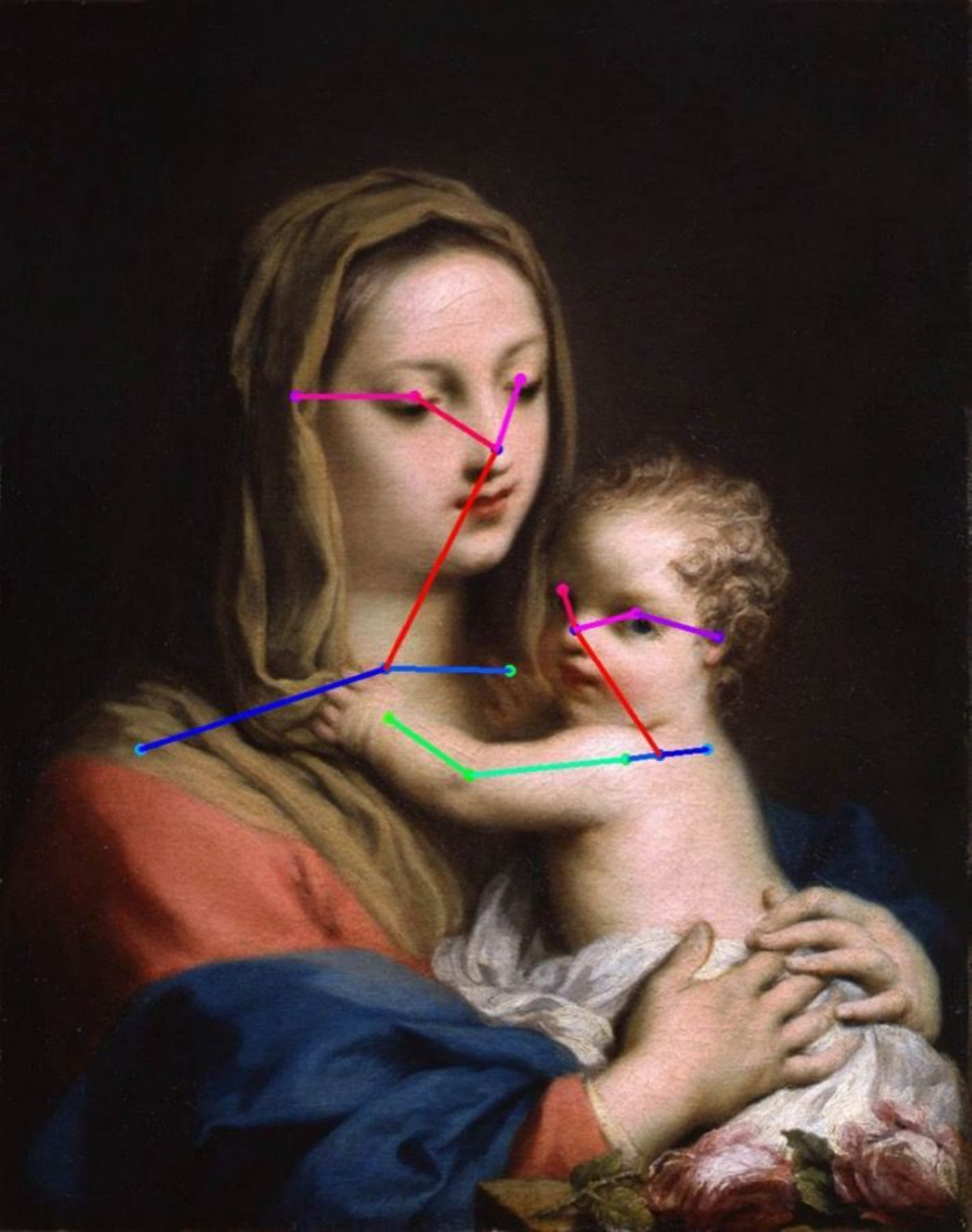}
			\caption{}
			\label{fig:maria-poseline-fix_a}
		\end{subfigure}
		\begin{subfigure}{0.2\textwidth}
			\includegraphics[width=\textwidth]{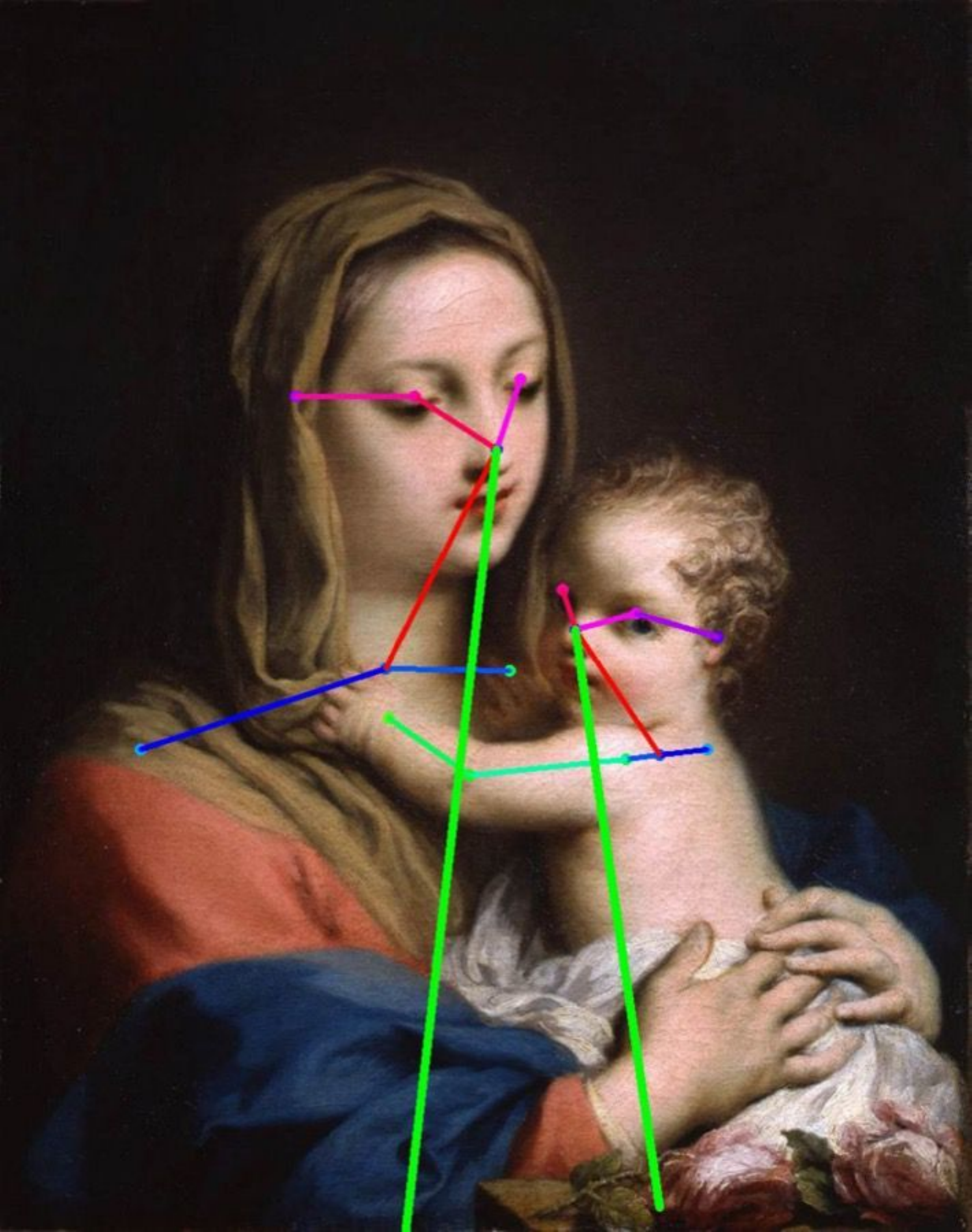}    
			\caption{}
			\label{fig:maria-poseline-fix_b}
		\end{subfigure}
		\begin{subfigure}{0.28\textwidth}
			\centering
			\includegraphics[width=\textwidth]{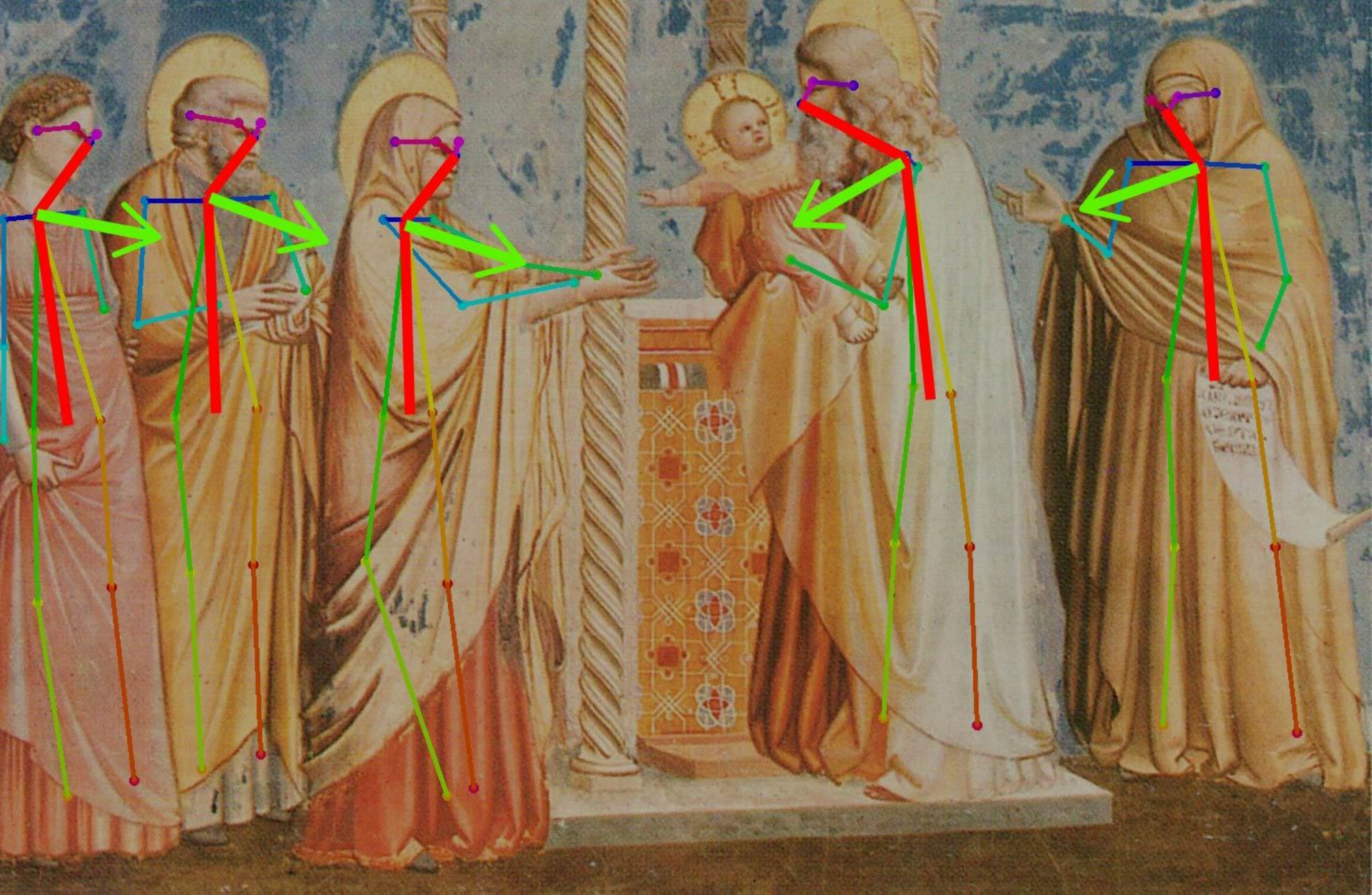}
			\caption{}
			\label{fig:bisection-example_a}
		\end{subfigure}
		\begin{subfigure}{0.28\textwidth}
			\centering
			\includegraphics[width=\textwidth]{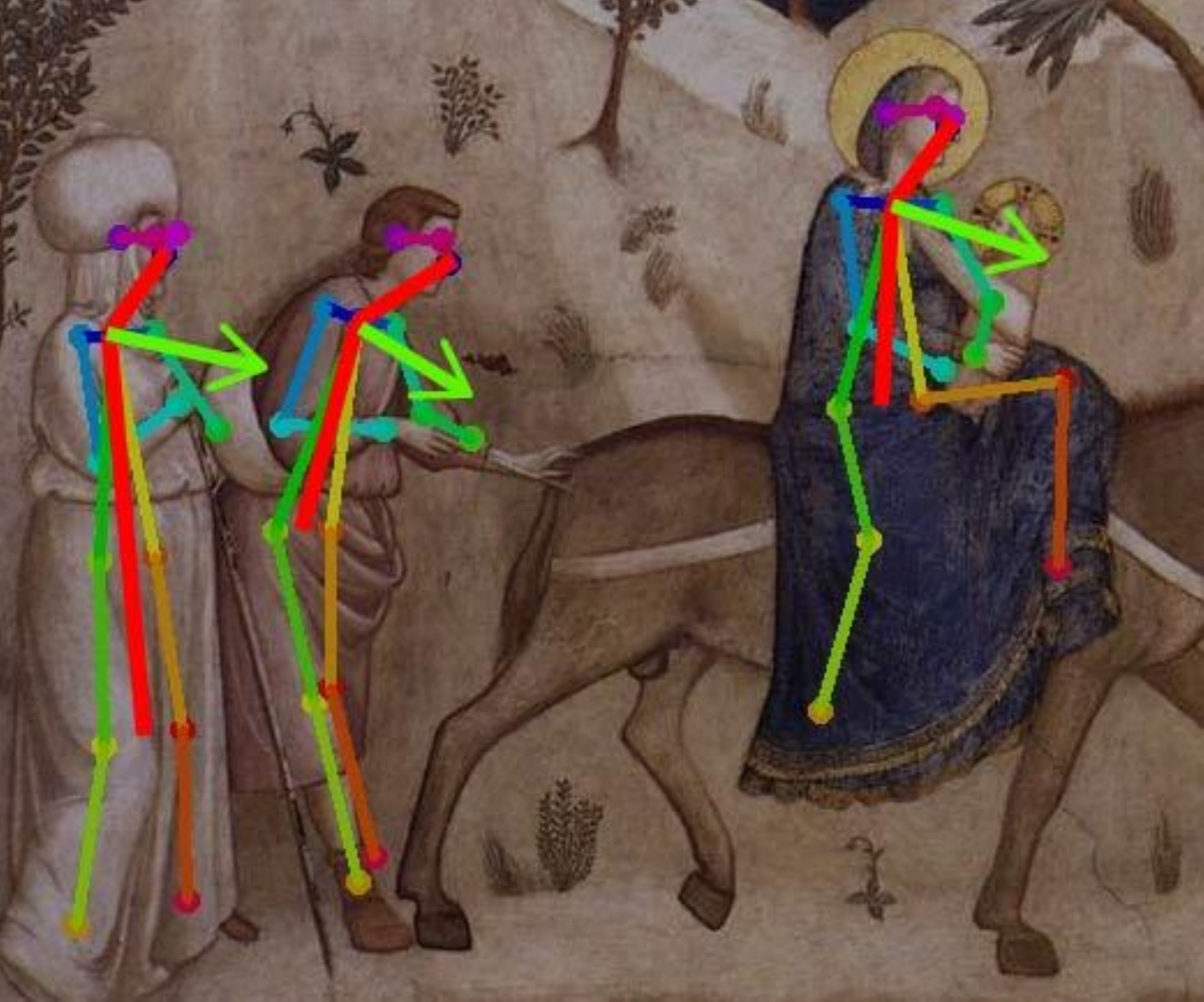}     
			\caption{}
			\label{fig:bisection-example_b}
		\end{subfigure}
		
		\caption{\subref{fig:maria-poseline-fix_a}-\subref{fig:maria-poseline-fix_b} A painting from the \textit{virgin and child} iconography; \subref{fig:maria-poseline-fix_a} all detected keypoints are visualized and no regular poseline generated because of the missing lower body part; \subref{fig:maria-poseline-fix_b} poseline fallback visualized in green; \subref{fig:bisection-example_a}-\subref{fig:bisection-example_b} Giotto's paintings with OpenPose detections; nose-neck and neck-mid-hip segments highlighted in bold red; bisection vector between them in bold green. Since this is a rough estimate, we can observe that it points slightly more downwards as compared to the correct body orientation which we correct with the correction angle $\rho$.
		}
		\label{fig:maria-poseline-fix}
	\end{figure*}
	
	\subsection{ICC++: Improvements over ICC}\label{subsec:ICC++}
	\icc is the improved version of the ICC method. The main goal is to design the features that would guide a composition-based image retrieval system. First, we provide corresponding improvements to the ICC algorithm for each drawback mentioned in \cref{subsec:iccdrawbacks}.
	
	\begin{enumerate}
		\item \label{subsubsec:posefallback} \textbf{Pose fallback:}	As described in ICC drawbacks (\cref{subsec:iccdrawbacks}, \cref{subsec:limit-triangle-abstraction}), ICC failed to generate poselines when lower body keypoints were missing. As a correction, we introduce `pose fallback' extrapolating the poseline by extending the neck-nose segment downward by three times its original length.  This issue (\cref{fig:maria-poseline-fix_a}) along with its corresponding fix (\cref{fig:maria-poseline-fix_b}) is shown in \textit{Mary} from the \textit{virgin and child} iconography.
		
		\item \label{subsubsec:updated-bisection} \textbf{Corrected Body Orientations:} To compute the global action centers, we additionally consider characters' body orientations as approximate gaze directions. Since the current SOTA gaze methods did not work on art historical images, we developed an approximation of the body orientation, using a bisection angle between the tuple of keypoints: (nose, neck, mid-hip). As observed in \cref{fig:bisection-example_a} and \cref{fig:bisection-example_b}, the bisection angle can be seen as an approximation of the body direction. 
		A small offset is caused due to this approximation that is fixed using a correction angle $\rho$.
		
		\item \textbf{Corrected Bisection angle:} To correct the bisection angles, we introduce new hyper-parameters that allow control of the cone shape and length, which helps mitigate this problem.
	\end{enumerate}

	\subsection{ICC++ Retrieval System}\label{subsec:ICC++ret}
\begin{figure}[!t]
		\centering
		\includegraphics[page=1,width=\linewidth]{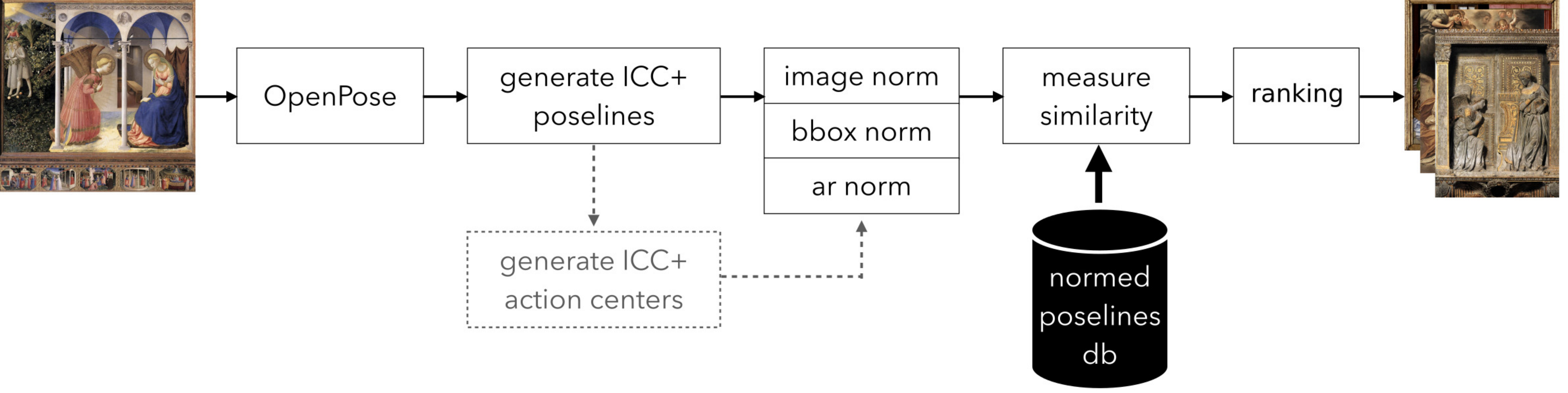}
		\caption{ICC++ retrieval pipeline.}
		\label{fig:ICC++retrieval-pipeline}
	\end{figure}

	In this section, we describe the retrieval system (\cref{fig:ICC++retrieval-pipeline}), discuss various normalization techniques, methods for comparing images with multiple poselines, and new retrieval methods in detail. 
	\icc uses the compositional elements as a measure of similarity for retrieval of similar images. 
	The pipeline is shown in \cref{fig:ICC++retrieval-pipeline}. 
	Given a query image, the algorithm generates poselines and action centers, then these poselines undergo a normalization technique as a pre-processing step for the similarity comparison.
	After normalization, the normalized poselines from the query image are compared with all pre-computed poselines from the target images in the database.  
	After comparison, the database images are ranked by the similarity measure and the top $N$ ranks are presented to the user.

	\subsubsection{Normalization of poselines}\label{subsubsec:norms} 
\begin{figure*}[!t]
		\centering
		\begin{subfigure}{0.48\textwidth}
			\centering
			\includegraphics[height=2cm]{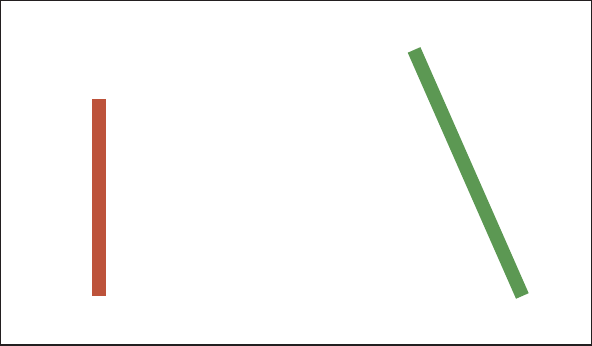}
			\caption{}
			\label{fig:imgsizenorm-problem_a}
		\end{subfigure}
		\begin{subfigure}{0.19\textwidth}
			\centering
			\includegraphics[height=2cm]{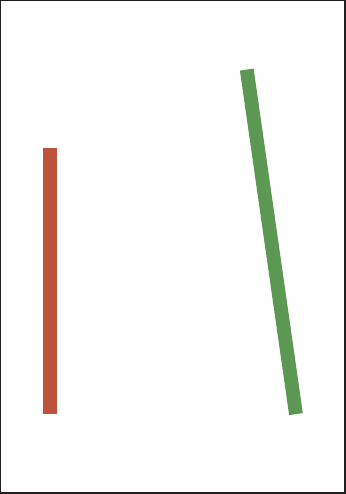}
			\caption{}
			\label{fig:imgsizenorm-problem_b}
		\end{subfigure}
		\begin{subfigure}{0.28\textwidth}
			\centering
			\includegraphics[height=2cm]{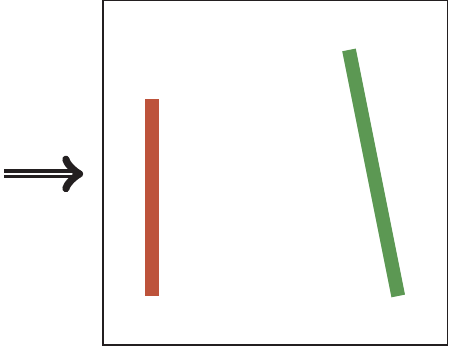}
			\caption{}
			\label{fig:imgsizenorm-problem_c}
		\end{subfigure}	
		\caption{Image normalization problem: \subref{fig:imgsizenorm-problem_a} and \subref{fig:imgsizenorm-problem_b} show two different composition diagrams containing two poselines each. While the compositions are quite different, the image min-max-normalized version \subref{fig:imgsizenorm-problem_c} is exactly the same for both.}
		\label{fig:imgsizenorm-problem}
	\end{figure*}

Normalizing the poselines is an important aspect for increasing the robustness of the similarity measurement. Therefore, to compare poselines across different images, they need to be normalized without compromising the compositional structure of the image. We aim to increase the robustness by bringing the poselines into a more generic representation by normalizing the top and bottom coordinate points of the poseline. One common approach for normalization is min-max normalization. We can compare images of different sizes by normalizing the poses with the image height and width (\texttt{image norm}). 
	By this method, however, we lose the aspect ratio information and the poselines are stretched or compressed along the $x$ or $y$-axis, as shown in \cref{fig:imgsizenorm-problem}. 
	This leads to high similarity scores even if the un-normalized image might have a different composition.  To reduce this problem, we present two new methods of normalizing the poselines using compositional elements: (a)~a keypoint bounding box normalization and (b) a global action region-based normalization.

	\paragraph{Keypoint bounding box normalization} (\texttt{bbox norm})
	We calculate the bounding box that contains all the characters and then normalize the keypoints using this bounding box. 
	By using this normalization, we aim towards a translation- and scale-invariant similarity measurement.
	However, this approach is very sensitive to missing pose detections as the minima and maxima values can vary drastically if the underlying pose-detector misses the detection of a single pose.

	\paragraph{Global action region based normalization} (\texttt{ar norm})
	Therefore, we propose a dynamic global action region based normalization.
	It is robust against the sub-picture problem and at the same time does not have the issues of image distortion as in \texttt{image norm}.
	We encode a piece of important composition information provided by the action regions into the poselines.
	For the normalization, we use the centroids of all global action regions. 
	We normalize all top-and-bottom poseline points by subtracting the global action centers. 
	In this way, the poselines become translation-invariant \wrt the image but not to the composition diagram, \ie the action centers themselves. 
	Since it is possible to have multiple action regions, we will have a set of normalized poselines, one normalized poseline set for each action region and hence the output is a set of sets of poselines. 
	Due to this change in the output format, we also have to change the similarity measurement between those normalized poselines. 
	
	\begin{figure*}[!t]
		\begin{subfigure}{0.24\textwidth}
			\centering
			\includegraphics[width=\textwidth]{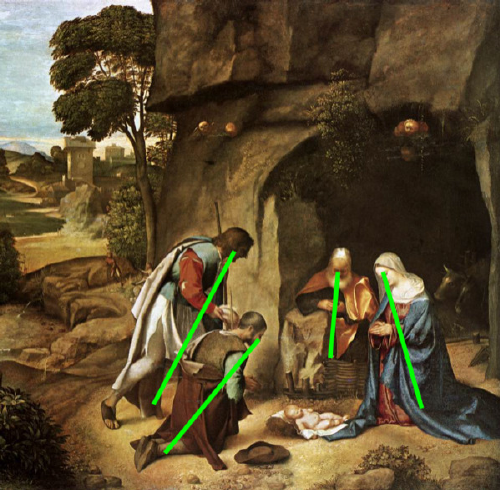}
			\caption{}
			\label{fig:bboxnorm-problem_a}
		\end{subfigure}
		\begin{subfigure}{0.24\textwidth}
			\centering
			\includegraphics[width=\textwidth]{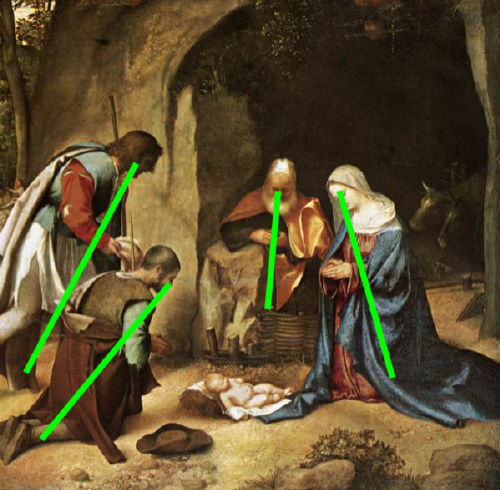}
			\caption{}
			\label{fig:bboxnorm-problem_b}
		\end{subfigure}	
		\begin{subfigure}{0.24\textwidth}
			\centering
			\includegraphics[width=\textwidth, page=2]{images/method/similarity/bboxnorm_problem_tikz_graphic.pdf}
			\caption{}
			\label{fig:bboxnorm-problem_c}	
		\end{subfigure}	
		\begin{subfigure}{0.24\textwidth}
			\centering
			\includegraphics[width=\textwidth, page=1]{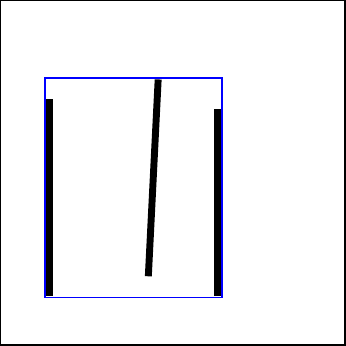}
			\caption{}
			\label{fig:bboxnorm-problem_d}
		\end{subfigure}
		\label{fig:bboxnorm-problem}
		\caption{Many images in the dataset are just a sub-image (like \cref{fig:bboxnorm-problem_b} of the full painting \cref{fig:bboxnorm-problem_a}). From the picture itself, it is not clear whether the picture covers the whole image or whether it is just a part of a bigger painting -- ideally, the comparison method should be robust against this. If the underlying pose detector fails to detect the pose needed for generating the red poseline in \cref{fig:bboxnorm-problem_c} a wrong bounding box would be generated (\cref{fig:bboxnorm-problem_d}). This shows that the output of the bounding-box-normalization is not robust to the sub-image problem.}
	\end{figure*}
	
	\subsubsection{Similarity measurement for multiple set of poselines}\label{subsubsec:similarity-multipose}
	For measuring the similarity between two images based on the poselines, we compare all the poselines from the query image $P_q$ to the target image $P_t$.
	
	\paragraph{Single poseline similarity} The similarity between two poselines is calculated by taking the mean of the Euclidean distance between their top and bottom points:
	
	\begin{equation}\label{eq:f-similarity-pl}
	f(p_q, p_t) = \frac{|p_q^{top}-p_t^{top}|+|p_q^{bottom}-p_t^{bottom}|}{2}; \;p_q \in P_q\,\text{,}\;p_t \in P_t
	\end{equation}
	\begin{figure*}[t]
		\centering
		\begin{subfigure}{0.45\textwidth}
			\centering
			\includegraphics[scale=0.6]{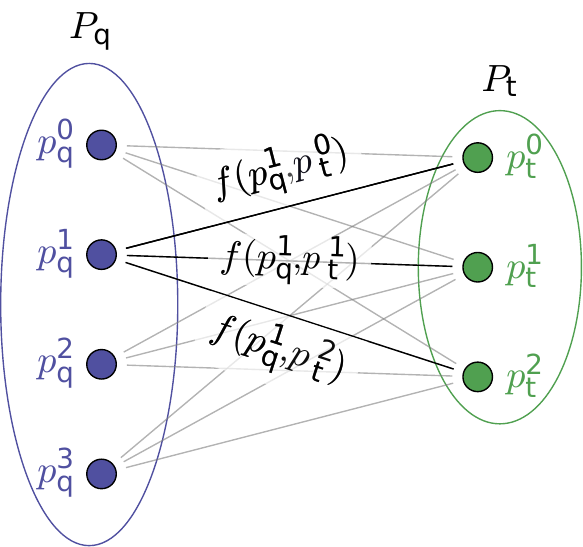}	
			\caption{}
			\label{fig:bipart-graph_a}
		\end{subfigure} 
		\begin{subfigure}{0.45\textwidth}
			\centering
			\includegraphics[scale=0.6]{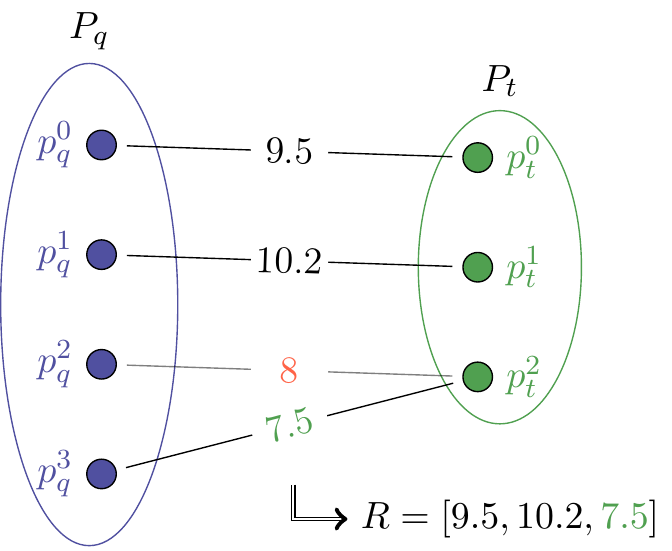}	
			\caption{}
			\label{fig:bipart-graph_b}
		\end{subfigure} 	
		\caption{\subref{fig:bipart-graph_a} Output of the first step of Algorithm 1 (Supplementary (Sec. 7.3)) is a fully connected bipartite graph of query image poselines $p_q^i \in P_q$ and target image poseline $p_t^i \in P_t$. All edges are weighted with the poseline distance $f(p_q^i,p_t^j)$, example edges for node $p_q^1$ have been highlighted; \subref{fig:bipart-graph_b} Illustration of the second step of Algorithm 1 (Supplementary (Sec. 7.3)). Each node is allowed to have only one edge. Since we iterate over a sorted list of edges, the lower distance in green is chosen over the red distance for node $p_t^2$.
		}
		\label{fig:bipart-graph}
	\end{figure*}
	
	\paragraph{Multi-poseline similarity -- bipartite graph generation} The similarity between two images (each defined by multiple poselines) is calculated by summarizing an approximation of a weighted bipartite minimum matching graph of poselines (\cref{fig:bipart-graph_b}).
	The bipartite graph has one node for each query poselines $p_q$ on the left, one node for each target poselines $p_t$ on the right, and connections between the left and right nodes are weighted by the single poseline similarity $f(p_q, p_t)$. 
	This approximated bipartite minimum matching graph is generated by creating a fully-connected bipartite graph (\cref{fig:bipart-graph_a}) and then applying our Algorithm 1 (Supplementary (Sec. 7.3)) to transform the fully-connected bipartite graph to an approximation of a bipartite minimum matching graph, with a maximum of one edge for each node (\cref{fig:bipart-graph_b}).

	\paragraph{Multi-poseline similarity -- bipartite graph summarization} The resulting list ($R$) of poseline distances is then filtered and summarized in three metrics: \texttt{hit ratio} ($r_\text{hr}$), \texttt{normalized mean distance} ($r_\text{nmd}$) and \texttt{combined ratio}($r_\text{cr}$) each describing the similarity between the two given images:
	\begin{align}\label{eq:summarize-weights}
		r_\text{hr} &= \frac{|R_f|}{\max\{|P_q|,|P_t|\}} \text{,} &&\text{ where } R_f = \{ d | d \in R \land d < \beta \}\\
	r_\text{nmd} &= \frac{\beta - r_\text{md}}{\beta} \text{,} &&\text{ where }  r_\text{md} = \frac{\sum_{d \in R_f} d}{|R_f|}\\
	r_\text{cr} &= r_\text{hr} * r_\text{nmd} 
	\end{align}
	where $d$ represents the edge weight and $\beta$ is used to filter out all outlier poseline pairs. $r_\text{hr}$ is a metric that measures the amount of matched poselines pairs and the values lie between 0 (worst match) and 1 (best match). The denominator for $r_\text{hr}$ prevents matching a target image consisting of several poselines with a query image with just a few poselines. 
	The reason is that the target image with a larger number of poselines is more likely to have a higher number of matching poselines with the query image when compared to a target image with fewer poselines.
The normalized mean distance of all matched poseline pairs is denoted as $r_\text{nmd}$. 
A value of $r_\text{nmd}=1$ means all matched poselines are perfectly aligned and a value close to zero means that even if the poselines match, the match is bad. Since both values ($r_\text{hr}$ and $r_\text{nmd}$) are in the range of 0 to 1, we can simply combine them to a combined ratio $r_\text{cr}$ to provide a single value describing the compositional similarity between two images. Various combinations of these three values $r_\text{hr}$, $r_\text{nmd}$, and $r_\text{cr}$ can then be used to rank the retrieved results (\cref{subsec:method-sorting}).
	
	\subsubsection{Combination with additional features.}\label{subsec:method-additional-combination}
	Optionally, our compositional features can be combined with other traditional or deep features to measure the similarity between two given images. Therefore, a similarity measurement for these additional features ($r_\text{a}$) has to be defined, which produces a single similarity value ranging from 0 (very similar) to n (not similar). This value can then be combined with our similarity measurement using the lexicographic sorting approach described in the next section or one of the following weighted combinations, where $w_a$ adjusts the balance between the additional features and the ICC++ features (if $r_\text{a} \in [0,1]$ then $w_a$ describes the contribution of $r_\text{cr}$ in percent). \cref{eq:combi-deep1} and \cref{eq:combi-deep2} describe the multiplicative and additive weighted ($w_a$) combination of the different features ($r_\text{a}$, $r_\text{cr}$) respectively:
	\begin{equation}\label{eq:combi-deep1}
	r_\text{combi1} = (r_\text{a} * (1-w_a)) * (1 - (r_\text{cr} * w_a))
	\end{equation}
	\begin{equation}\label{eq:combi-deep2}
	r_\text{combi2} = (r_\text{a} * (1-w_a)) + (1 - (r_\text{cr} * w_a))
	\end{equation}
	
\subsubsection{Retrieval and ranking of results.}\label{subsec:method-sorting}
We use the similarities to rank the results for the retrieval task. 
We first calculate the action regions and poselines for the query image and then iteratively calculate all the similarity values between the query image and all pre-computed action regions and poselines of the target images. 
	For each iteration, we store the query image, the target image and all the similarity values for this combination ($r_\text{cr}$, $r_\text{hr}$, $r_\text{nmd}$). If we also use the optional combination of the \icc method with other deep or traditional methods, we will have the following values $(r_\text{cr}, r_\text{hr}, r_\text{nmd}, r_\text{combi1},r_\text{combi2},r_\text{a},I_q,I_t)$, $I_q$ and $I_t$ are query and target images respectively. We then sort the values to obtain the final ranking of the retrieval results. The used sorting methods and their corresponding keys are shown in Supplementary (Tab. 9).
	
\section{Experiment Design}\label{sec:expsetup}
To evaluate the effectiveness of \icc, we designed a retrieval experiment and test it on our newly created dataset \textit{WGA500}. In this section, we introduce the new dataset called Web Gallery of Art 500 (WGA500) (\cref{subsec:datasets}), describe the different experimental setups with traditional computer vision and deep learning methods with our \icc method, and also compare \icc with the state-of-the-art (SOTA) LATP method quantitatively and qualitatively (\cref{subsec:Setup}). Since there are no labeled datasets with composition labels, the retrieval setup quantifies the compositional similarity between images based on image-level labels. 
	
	\subsection{Web Gallery of Art 500 (WGA500) dataset}\label{subsec:datasets}
	There is no dataset publicly available that matches our method for the task of retrieval based on image compositions. Hence, we curated a balanced dataset with 500 randomly chosen images of 5 iconographies (100 per each iconography) from the publicly available WGA~\cite{danielmarxWebGalleryArt} dataset of art historic paintings. Since our focus is to understand the compositions within iconographies, we considered ``adoration'', ``annunciation'', ``baptism'', ``nativity'' and ``virgin and child'' as underlying iconographies. Each iconography name acts as the class label for an image in the retrieval pipeline. Each of these iconographies has a distinct compositional structure, making them viable data to analyze our method. Accordingly, it is more likely to have the same composition occurring multiple times within the same iconography class than across multiple iconographies. Even so, images in a given iconography are also quite distinct of styles, author, period, and general structure. As a result, very few images within each class (iconography) have the exact same composition. 
	
	
	\subsection{Experimental Setup}\label{subsec:Setup}

	\subsubsection{Evaluation Metrics}
	We calculate the precision P@k and recall R@k metrics for the retrieval experiments at all \textit{k}-th ranks of each retrieval result ($k \in [1, 2, \dots, 10]$). As a final reported evaluation metric, we use the mean of the precision (mP@k) and recall (mR@k) values over all the 500 query-images. We compare our method with traditional features, deep features and the LATP~\cite{jenicekLinkingArtHuman2019} approach using mP@1, mP@2, mP@5 (as presented in their paper), along with mAP. In the next section, we describe model selection and hyperparameter settings for \icc, and then explain the setup of the traditional and deep features.
	
	\subsubsection{Evaluation Protocol}
	Before comparing \icc with other approaches, we conduct an ablation study to setup the baseline of our method, \ie we find the best working pipeline. 
	We presented various sub-methods (improvements to ICC) that can be interchanged and combined simultaneously (poseline fallback and bisection fallback in \cref{subsec:iccdrawbacks}, cone base scaling in Supplementary (Sec. 7.2.1), and various normalization methods in \cref{subsubsec:norms}). 
	Due to the wide variety of methods and hyperparameter combinations, we conduct a two-step cross-validation procedure to find the best model for the dataset:
	
    (1)~We compare all the interchangeable \icc method options to find a single \icc baseline method. For the baseline method, we have deactivated all additional ICC++ features. The baseline method, therefore, consists of the following constraints: 
    \begin{enumerate*}[label=(\textit{\alph*})]
        \item No normalization of the poselines, 
        \item \texttt{cr\_desc} is used as the result set sorting method (see Supplementary (Sec. 7.5) for a definition of all sorting methods),
        \item all fallback methods turned off,
        \item filter-threshold $\beta = 150px$, 
        \item all the cone parameters from the original ICC method were used, 
        \item correction angle $\rho = 20^{\circ}$,
        \item cone opening angle $\omega = 80^{\circ}$, 
        \item cone scale factor $\sigma = 10$, and 
        \item cone base scale factor $\eta = 0$. 
    \end{enumerate*}

	We then introduce a single method one at a time, starting with all normalization methods, followed by changing the sorting methods, and lastly enabling the poseline fallback. For all min-max-based normalization techniques, we changed the threshold to $0.15$ since this is synonymous to the $150px$, because all images in the dataset have been up- or down-scaled to exactly $1000px$ ($150px/1000px = 0.15$). Afterwards, we changed the baseline method to use \texttt{ar norm} instead of no normalization and evaluate all method options that only affect the action regions.
	
(2)~We perform a grid search to find the best method for all possible combinations of predefined hyperparameter ranges, which is then compared with all the traditional and deep learning based methods.
	
	\subsubsection{Traditional CV methods}\label{subsubsec:tradmethods} As traditional features, we evaluated SIFT~\cite{ng2003sift}, BRIEF~\cite{calonder2011brief}, and ORB~\cite{rublee2011orb} features. 
	We used the OpenCV [39] implementation with standard parameters and grayscale images as input for all methods. 
	SIFT is the most common baseline method for traditional feature-matching algorithms. 
	BRIEF and ORB are efficient binary descriptors needing little storage and are therefore the only other features that are almost as small as our features in terms of storage size. 
	\textbf{SIFT.} For every image in the dataset, we precomputed 500 SIFT feature descriptors. 
    We then used knn-algorithm ($k=2$) for measuring the similarity between the two image descriptors. 
    After matching, a ratio test with a ratio of $0.75$ is performed as proposed in the SIFT paper \cite{loweDistinctiveImageFeatures2004}. 
    \textbf{BRIEF.} For detecting the feature-keypoints, we used the OpenCV STAR detector, a derivative of the CenSurE method \cite{agrawalCenSurECenterSurround2008}.
    We then evaluated the number of matching descriptors using a brute force matching algorithm with hamming distance measurement and cross-checking enabled to measure the similarity.
    \textbf{ORB.} The entire set up is the same as of BRIEF, but the only difference is the use of the OpenCV ORB detector for finding the descriptor-keypoints.
    For all these three methods, after matching and ratio-test (if conducted), as a final similarity value, we have either used the amount of filtered features or the amount divided by the maximum of available features in query and target image.
    
    In addition, we also generated global feature encodings on all of the local features using vector of locally aggregated descriptors (VLAD) ~\cite{jegou2010aggregating}. 
    In order to take care of the overfitting problem, we generated 10 clusters using hundred random images from the dataset and then later assigned the cluster to the rest of the images. 
    The assignment and evaluation method is followed as mentioned in \cite{jegou2010aggregating}.

	\subsubsection{Deep learning methods}\label{subsubsec:deepmethods} We evaluated VGG19~\cite{simonyanVeryDeepConvolutional2015} with batch normalization~\cite{ioffeBatchNormalizationAccelerating2015} pre-trained on the ImageNet. Additionally, we evaluated ResNet50~\cite{he2016deep} pre-trained on the ImageNet dataset and fine-tuned on the Places365 dataset~\cite{zhouPlaces10Million2017}. 
For both methods, we first precomputed the feature vectors for all images as with the other methods. During retrieval, we use the Euclidean distance ($r_\text{vgg1}$, $r_\text{resnet1}$) or the negative cosine similarity ($r_\text{vgg2}$, $r_\text{resnet2}$) to measure the similarity between two images.
	
  \subsubsection{Linking Art through Human Poses} We also compare our method with LATP~\cite{jenicekLinkingArtHuman2019}, the closest related work to ours. They define two similarity measurements to rank their results. Both are based on measuring the Euclidean distance between a pair of neck-point-normalized versions of the detected human poses.  They also implement a robust verification step (removing outliers) using the RANSAC algorithm. We re-implemented their work, because their code is not publicly available, and compared their results with ours on WGA500 dataset.
  
\section{Results}\label{sec:results}
	\subsection{ICC++ model selection and hyperparameter tuning}
	
	\begin{table}[t]
		\caption{Comparison of \icc baseline \vs all other method options. The first column is the main method and the second column its corresponding option; \textbf{mP@1} is mean precision value at 1 and \textbf{diff.} is the difference (in m@P1) of a given method with the \texttt{baseline} method.}
		\label{table:eval-step1-baseline}  
		\centering
		\begin{tabular}{@{}l|lcc@{}}
			\toprule
			\textbf{Method} & \multicolumn{1}{c}{\textbf{Option}} & \textbf{mP@1} & \textbf{diff.}  \\ \midrule
			\texttt{baseline} & \multicolumn{1}{c}{\texttt{-}} & 37.6 & - \\
			\midrule
			\texttt{norm} & \texttt{(1) image norm} & 38.6 & \greenten 1.0 \\
			& \texttt{(2) bbox norm} & 34.6 & \redten -3.0  \\
			& \texttt{(3) ar norm} & 39.2 & \greenten 1.60 \\ 
			\midrule
			\texttt{fallback} & \texttt{(1) poseline} & \textbf{41.2} & \greenthirty 3.6 \\
			& \texttt{(2) bisection} & 37.4 & \redten -0.2 \\
			\bottomrule
		\end{tabular}
		
	\end{table}
	
	\subsubsection{Step 1: ICC++ (baseline vs. other method options)}\label{subsubsec:gridsearch1}
	First we perform a two-step evaluation (\cf \cref{sec:expsetup}) of our method. In \cref{table:eval-step1-baseline}, we can see the first step results where we compare each \icc method option to the baseline. The \icc baseline is the one with none of the method options active. To evaluate the best option, we use the mean precision value at 1 (\textbf{mP@1}). The difference between the baseline model and any method option is denoted as \textbf{diff.}. 
	
\Cref{table:eval-step1-baseline} shows that normalizing (\texttt{norm} method) the input poselines (\cf \cref{subsubsec:norms}) improves the retrieval results in some cases, especially the \texttt{ar norm}. 
While the \texttt{image norm} option slightly improves, the \texttt{bbox norm} option does not show any improvement because it is sensitive to missing poses. 
The newly introduced method of \texttt{fallback} with \texttt{poseline} option logged the highest improvement of \SI{3.60}{\percent}. 
This is due to the fact that we have significantly more features and fewer poselines were missed. 
The \texttt{fallback} method allowes the detection of \num{2736} poses as compared to \num{2098} poses without fallback for the entire dataset. 
Conversely, the use of \texttt{bisection} option deteriorates the result.
	
	
	We see a decrease in performance when we combine the \texttt{ar norm} option of the normalization method with the \texttt{poseline} option of the fallback method (mP@1 40.6) in comparison to just the \texttt{poseline} option (mP@1 41.2). 
	A possible explanation for this could be that the \texttt{ar norm} is dependent on good action regions and therefore its precision depends on a good selection of the cone hyper-parameters. 
	
	\subsubsection{Step 2: Hyperparameter Tuning}\label{subsubsec:hyperparms}
	We found that the best performance of the method was when we combined \texttt{poseline} option of the fallback method (\cref{table:eval-step1-baseline}) with \texttt{ar norm} option of the normalization method resulting in an \texttt{mP@1} of $41.2$. We call this method as \icc U (U:\ untuned) because the network itself has not been altered, \ie fine-tuned.
	To validate the hyperparameter selection, we perform a grid-search for the \texttt{ar norm} and all of the cone hyper-parameters (Supplementary (Sec 7.4)) 
	for analyzing the high dependency between the poseline fallback and the other normalization methods. 
	Since this method is tuned on the dataset, we call it \icc T (T:\ tuned).
	
	\begin{table}[t]
		\centering
		\caption{\icc \vs Low-level features}				
			\begin{tabular}[t]{lcccc}
				\toprule
				\textbf{Method} & \texttt{ mP@1} & \texttt{ mP@2} & \texttt{ mP@5} & \texttt{mAP} \\
				\midrule
				SIFT$_{VLAD}$ & \textbf{48.0}  &  38.0   &    27.0    &   23.0  \\        
				BRIEF$_{VLAD}$ &  47.0  &  38.0   &    29.0    &   23.0  \\        
				ORB$_{VLAD}$ &  43.0  &  35.0   &    26.0    &   22.0  \\        
				\midrule
				ICC  &   38.00  &   35.60   &   31.68 &  24.5 \\
				\icc U &   41.20          &       \textbf{40.20}   &       \textbf{38.04} & \textbf{30.9} \\            
				\midrule
				\midrule
				\icc U $\bigoplus$ SIFT$_{VLAD}$   & 42.4  & 40.2 & 36.5 &  26.4 \\
				\icc U $\bigoplus$ BRIEF$_{VLAD}$  & 42.0 & 40.6 & 36.6 &  26.4 \\
				\icc U $\bigoplus$  ORB$_{VLAD}$   & 42.4  & 40.4 & 36.9 &  26.5 \\
				\bottomrule
				\multicolumn{5}{l}{$\bigoplus$: concat operation}\\
			\end{tabular}                                                                                 
			\label{table:eval-traditional}	
	\end{table}

	\subsection{ICC++ vs. traditional image features}\label{subsec:tradcompare}
	As observed in \cref{table:eval-traditional}, the global VLAD embeddings of low-level image features are insufficient to retrieve results with the same iconography and compositions. 
	One possible reason could be that the global features generated using methods are heavily dependent on the local features and the cluster hyper-parameters, which can be improved by integrating spatial verification.
	This can be considered as a whole new experiment with the tuning of various hyper-parameters as hence we consider as one of our future works.
    However for art historical datasets, the intra-class distance can vary greatly and thus affect the low-level similarity.
    Our \icc method that uses a combination of low- and high-level features outperforms all traditional low-level methods.
    A combination of \icc with the low-level VLAD features improves the $mP@1$ and $mP@2$ while \icc still maintains its superior performance for $mP@5$ and $mAP$, showing consistently better performance. 
    From \cref{table:eval-traditional}, we can observe that only mP@1 for SIFT outperforms our method, while our method outperforms all of these global embeddings in all metrics. 
    We also present the results of SIFT, BRIEF and ORB local features and their combination with our features in Supplementary (Tab. 7). 
    The corresponding sorting algorithm used for each of the methods and comparison with the fine-tuned version of \icc (\icc T) is shown in Supplementary (Tab. 7). 
	
	\begin{table}[t]
	    \centering
	    \caption{\icc vs. Deep methods}
	    \begin{tabular}{lcccc}
					\toprule
					\textbf{Method} & \texttt{ mP@1}& \texttt{ mP@2} & \texttt{ mP@5} & \texttt{mAP} \\
					\midrule
					VGG19     &  58.80  & 56.90 &  52.36  & 31.3  \\
					ResNet50  &  55.00  & 50.70 &  46.76  & 29.6  \\
					\midrule
					ICC    & 38.00  & 35.60 & 31.68 & 24.5 \\
					\icc U & 41.20 &  40.20 & 38.04 & 30.9\\            
					\midrule
					\icc U $\bigoplus$ VGG19    & \textbf{63.40}  & \textbf{57.80}  & \textbf{53.44}  & \textbf{32.5} \\
					\icc U $\bigoplus$ ResNet50 & 58.8  & 54.2 & 49.4  & 30.8 \\
					
					\bottomrule
					\multicolumn{5}{l}{$\bigoplus$: concat operation}\\
				\end{tabular}                                           
				\label{table:eval-deep}     
	    
	\end{table}

	\subsection{ICC++ vs. Deep features}
	The deep methods use multiple convolutional layers which can identify and group the lower-level features for higher-level interpretations. Due to this, deep methods can measure the similarity between two images at a higher level compared to the traditional methods. Since the visual semantic style between the images in our dataset is similar for the same iconography, the deep methods outperform our method in the retrieval task. 
	
However, it is visually difficult to interpret whether the deep method outperforms \icc because the images of the same iconography have the same textures and style, or whether it is because they share a similar composition. In contrast, our method shows two clear advantages over deep methods: 
	\begin{enumerate*}[label=(\arabic*)]
		\item\icc features are low-dimensional and intelligible enough to be easily visualized on the image so that one can visually interpret the compositional elements used to retrieve them; 
		\item \icc features can be explained well based on the poselines and action regions compared to the high-dimensional complex deep features. 
	\end{enumerate*}
	
	In addition, we combine two different deep learning-based features with our \icc.
	The results (\cref{table:eval-deep}) show that this combination gives \SI{\approx 5}{\percent} increase in top-1 accuracy (\texttt{mP@1}) with both the deep methods (VGG19 and ResNet50). This highlights the importance of our features as a valuable addition to the feature representation that achieves better performance than deep-learning-only features. We discuss the corresponding visual results in detail in \cref{subsec:discussinterpret}. The sorting algorithm used for each method and comparison with the fine-tuned version of \icc (\icc T) is shown in Supplementary (Tab. 8).
	

	\begin{figure}[!h]
		\begin{subfigure}{0.24\textwidth}
			\centering
			\includegraphics[width=1\linewidth]{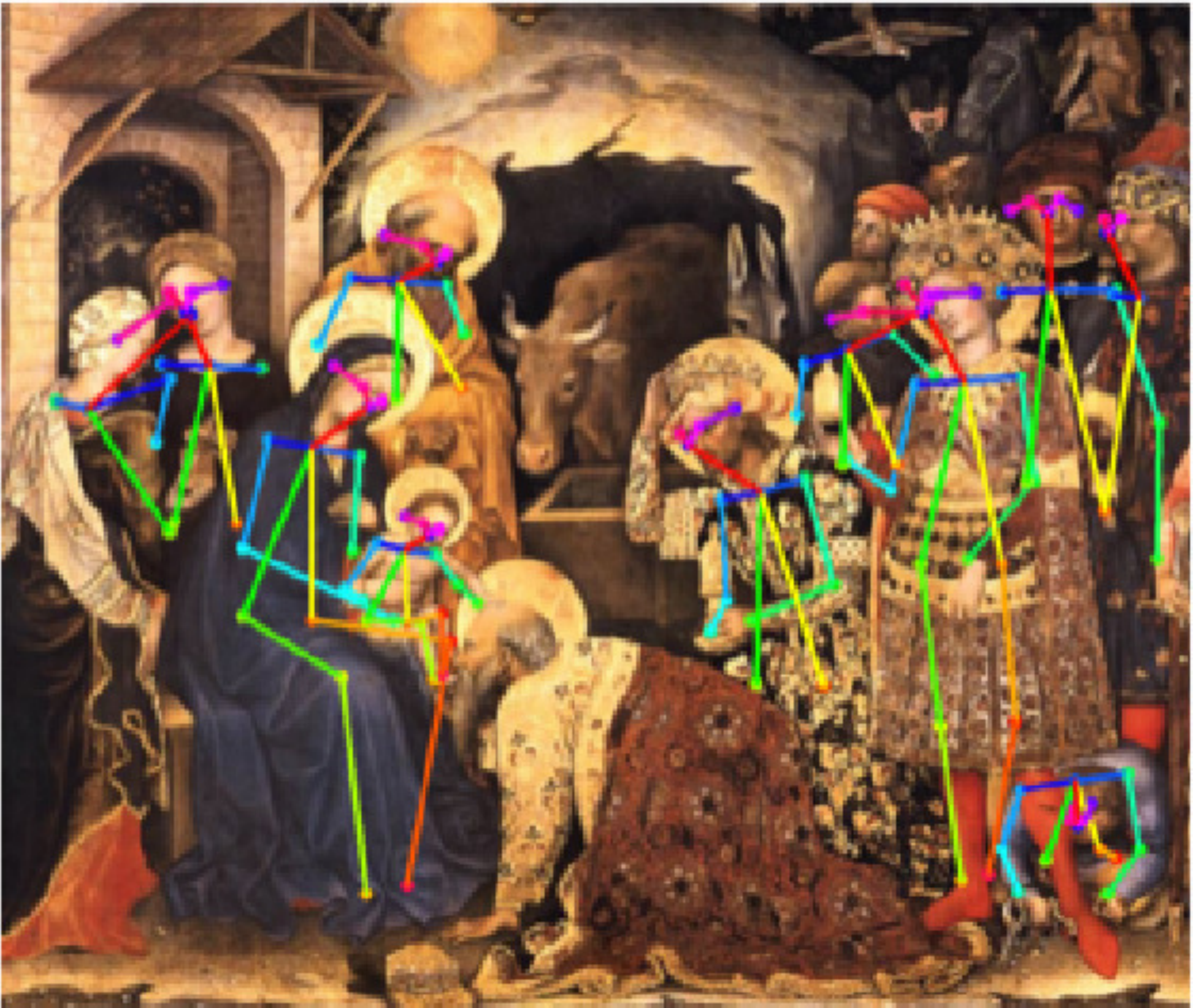}
			\caption{Query Image}
			\label{fig:eval:latp_robust_verify_1}
		\end{subfigure}
		\begin{subfigure}{0.24\textwidth}
			\centering
			\includegraphics[width=1\linewidth]{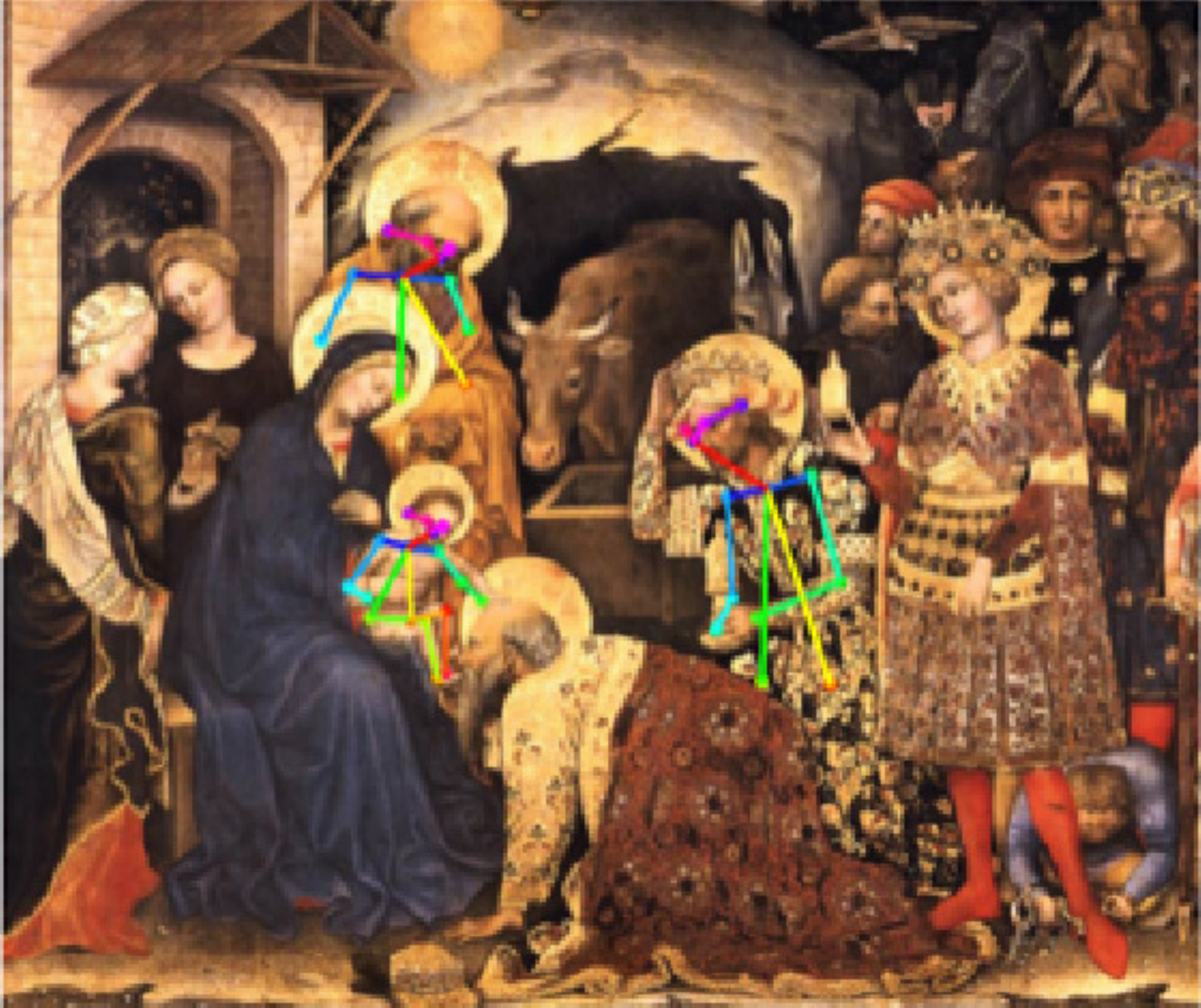}
			\caption{\textit{Matched Poses}}
			\label{fig:eval:latp_robust_verify_2}
		\end{subfigure}	
		\begin{subfigure}{0.24\textwidth}
			\centering
			\includegraphics[width=1\linewidth]{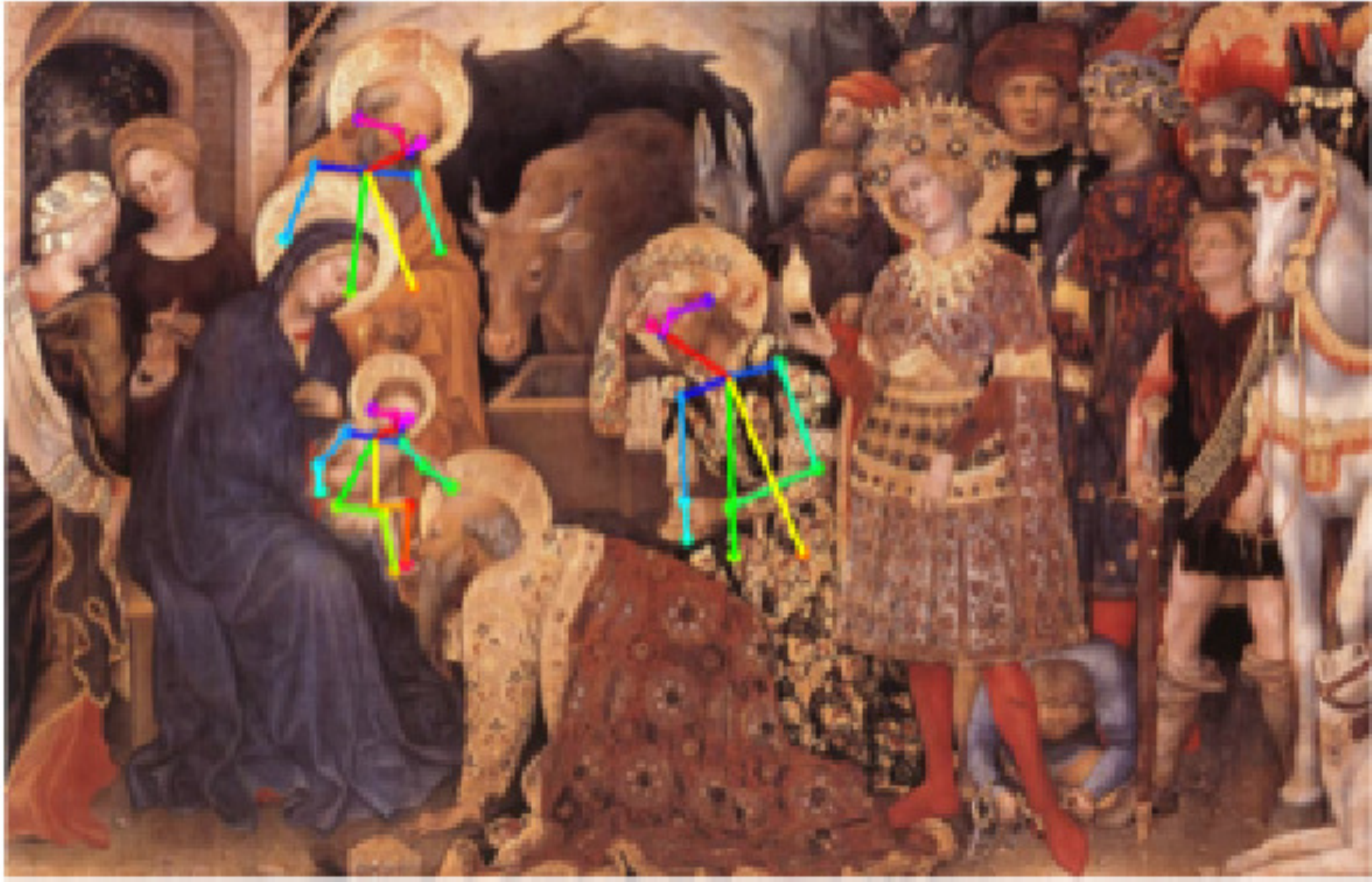}
			\caption{\textit{Matched Poses}}
			\label{fig:eval:latp_robust_verify_3}	
		\end{subfigure}	
		\begin{subfigure}{0.24\textwidth}
			\centering
			\includegraphics[width=1\linewidth]{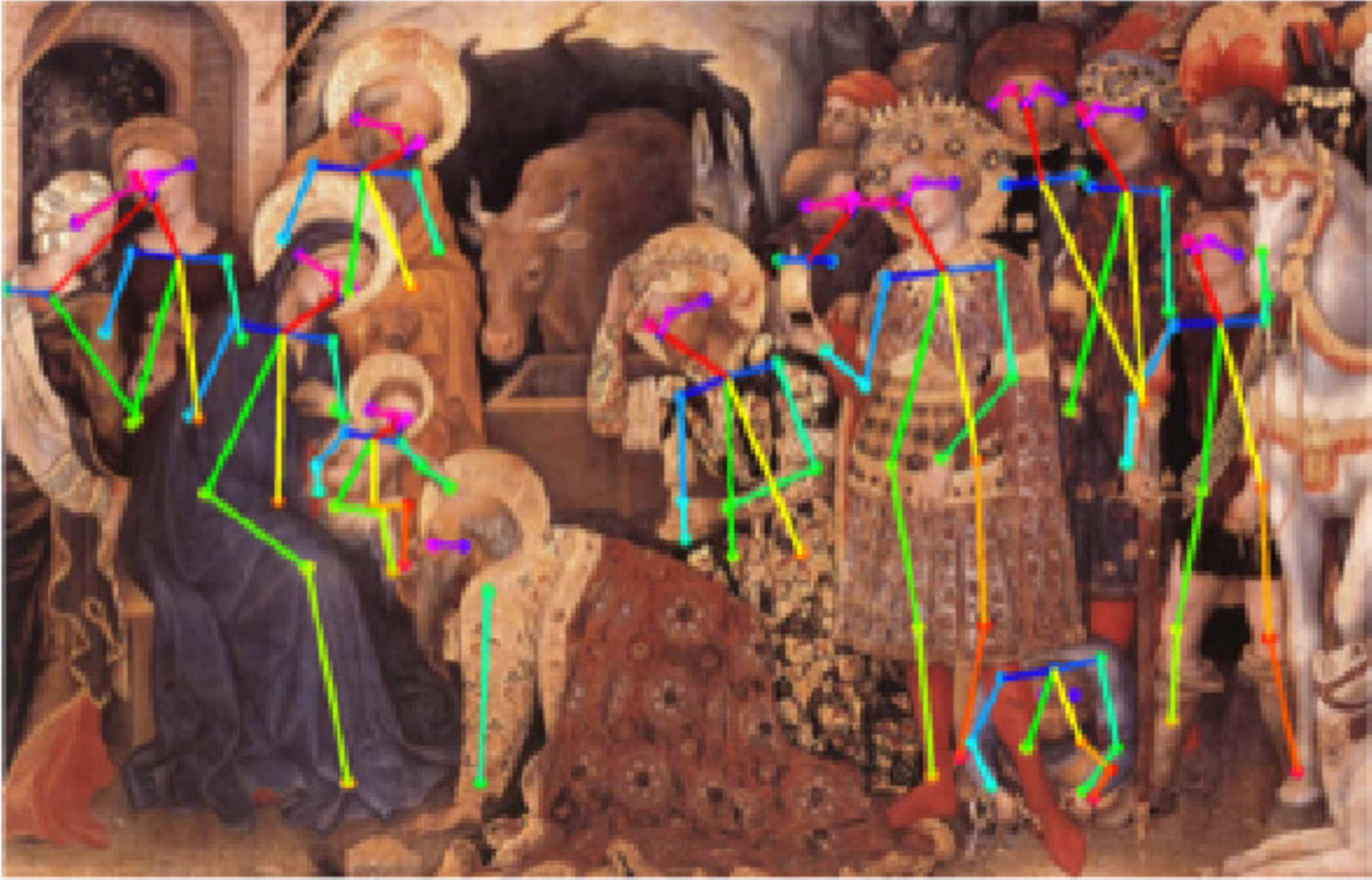}
			\caption{Target Image}
			\label{fig:eval:latp_robust_verify_4}
		\end{subfigure}
		\caption{Query image \subref{fig:eval:latp_robust_verify_1} is an image  from \textit{virgin and child} class. We can note that the retrieved target image \subref{fig:eval:latp_robust_verify_4} is a duplicate of the query image. \subref{fig:eval:latp_robust_verify_2} and \subref{fig:eval:latp_robust_verify_3} depicts \textit{matched poses} between the two images.}
		\label{fig:eval:latp_robust_verify}
	\end{figure}


	\subsection{ICC++ vs Linking Art through Human Poses}
	We can see in \cref{table:eval-latp}~that the vanilla LATP does not even outperform our untuned method (\icc U). 
	Using LATP with robust verification successfully detects copies of the same composition as seen in \cref{fig:eval:latp_robust_verify}. 
	But at the same time, it does not generalize the linking of images of the same iconography when the composition of the query changes slightly. 
	A possible reason could be that LATP lacks high-level composition abstraction. 
	Moreover, the robust verification worsens the results, \cf \cref{table:eval-latp} row 1 and 2 \vs row 3 and 4.
	
	In addition, we combine LATP with deep features to have a fair comparison with our method. The results (\cref{table:eval-latp}) show that our method (\icc U) combined with deep features outperform every combination of LATP and deep features. 
	
	\begin{table}[t]
		\caption{Top: \icc untuned (U) in comparison to LATP with or without robust verification and different distance matching methods; Bottom: Comparison of deep features combined with LATP and \icc untuned (U)}
		\begin{center}
			\begin{tabular}{lcccc}
				\toprule
				\textbf{Method} & \texttt{ mP@1}& \texttt{ mP@2}& \texttt{ mP@5}& \texttt{ mAP}\\ 
				\midrule
				LATP $dist_{t}$ (bipart distance), robust verify  &  27.4 &  27.9  &  26.2  & 27.6 \\
				LATP $dist_{min}$ (min distance), robust verify   & 29.6 &  27.6  &  27.2  & 23.4 \\
				LATP $dist_{t}$ (bipart distance)           &  29.2  &  30.4  &  27.9  & 27.9 \\
				LATP  $dist_{min}$ (min distance)          &  31.4  &  31.8  &  29.5  & 23.7 \\
				\midrule
				\icc U (Ours)    &   \textbf{41.2} & \textbf{40.2} & \textbf{38.4} &  \textbf{30.9} \\           
				\midrule
				\midrule
				LATP $dist_{t}$ $\bigoplus$ Resnet50 & 55.2 & 49.6 & 45.5 & 29.0 \\
				LATP $dist_{min}$ $\bigoplus$ Resnet50 & 54.8 & 49.8 & 46.4 & 29.1 \\
				\icc U $\bigoplus$ ResNet50 (Ours) & \textbf{58.8}  & \textbf{54.2} & \textbf{49.4}  & \textbf{30.8} \\
				\midrule
				LATP $dist_{t}$ $\bigoplus$ VGG19 & 59.0 & 57.0 & 52.3 & 31.4 \\
				LATP $dist_{min}$ $\bigoplus$ VGG19 & 59.4 & 55.9 & 51.4 & 30.6 \\
				\icc U $\bigoplus$ VGG19 (Ours)   & \textbf{63.40}  & \textbf{57.80}  & \textbf{53.44}  & \textbf{32.5} \\
				\bottomrule
				\multicolumn{5}{l}{$\bigoplus$: concat operation}\\
			\end{tabular}
			\label{table:eval-latp}
		\end{center}
	\end{table}

	\subsection{Discussion and Interpretations} \label{subsec:discussinterpret}
	In our qualitative evaluation, we discuss the best methods across each category (low-level, deep, and \icc), \ie SIFT, VGG19, \icc U, \icc U $\bigoplus$ VGG19, and LATP $dist_{min}$. We use the annunciation scene as it has a very clear and dominant composition with slight variation across all paintings inside the iconography. 
	
\begin{figure}[t]
		\centering
		\includegraphics[width=1\linewidth]{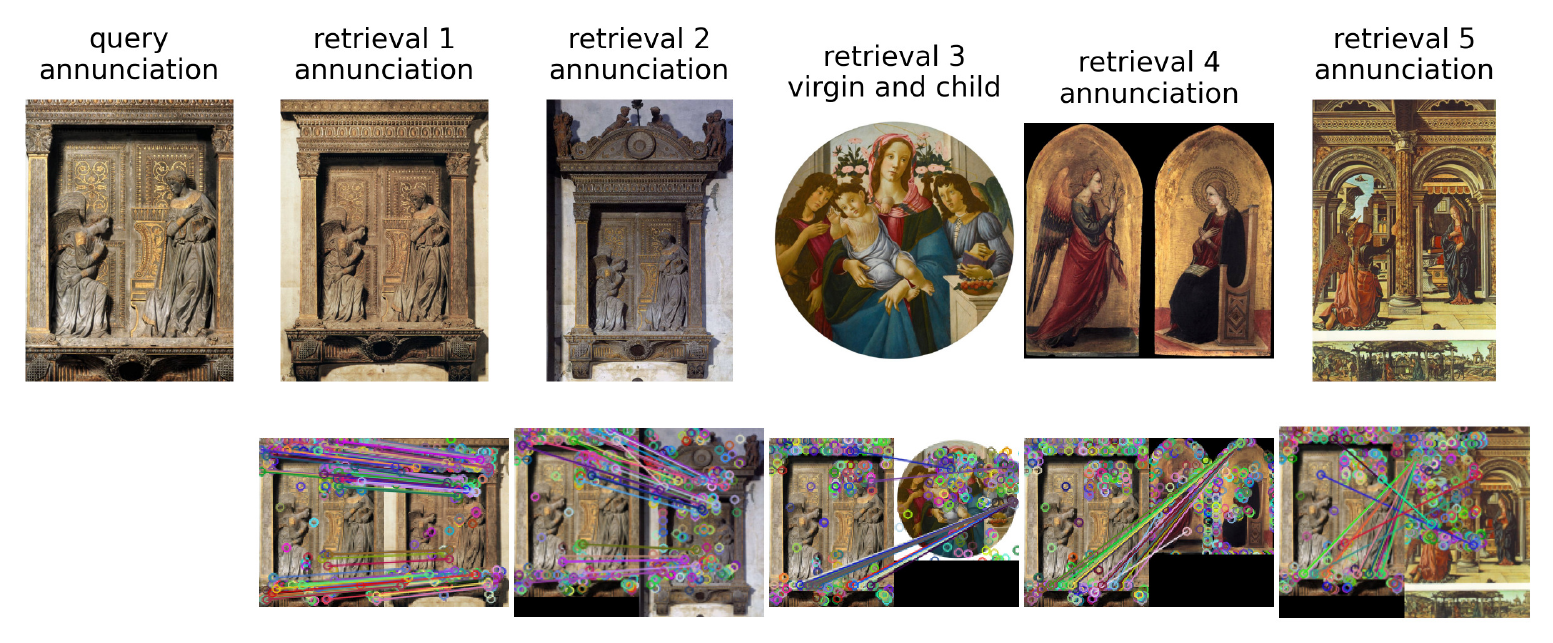}
		\caption{\textbf{SIFT results}: Top row: Top-5 retrieval results; bottom row: SIFT correspondences between query and target image.}
		\label{fig:eval:plots_sift_all}
	\end{figure}

	\textbf{SIFT} retrieves (almost) replicas or copies of the query in our dataset as we can observe in \cref{fig:eval:plots_sift_all}. If there are exact copies in the database, these low-level features could be sufficient to find matches within the dataset. On close observation, we can see in retrieval 3 to 5 of \cref{fig:eval:plots_sift_all} that the amount of matched features is relatively low. SIFT, therefore, is unable to match compositionally similar images. 
	
\begin{figure}[t]
		\centering
		\begin{subfigure}{0.95\textwidth}
			\includegraphics[width=1\linewidth]{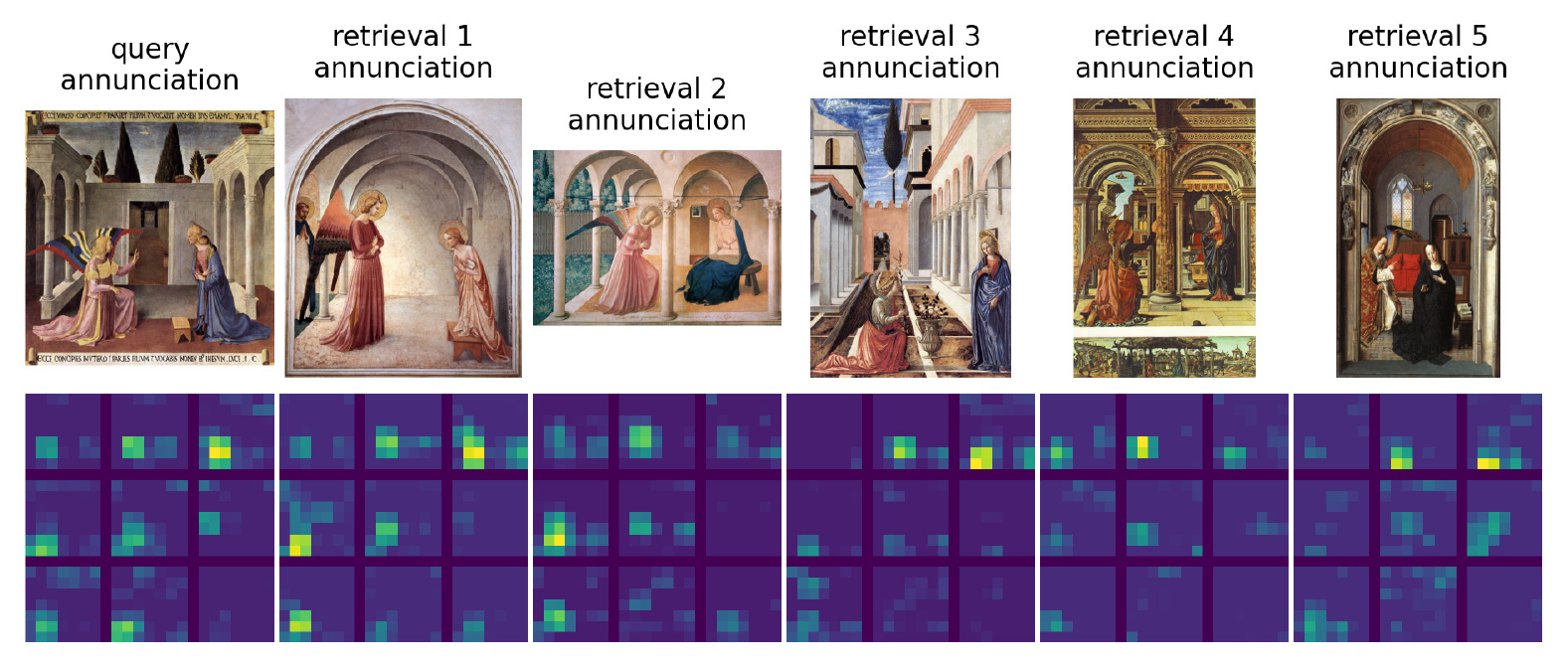}
			\subcaption{VGG19 (best)}
			\label{fig:eval:plots_vgg19_all}
		\end{subfigure}%
        \\
		\begin{subfigure}{0.95\textwidth}
			\includegraphics[width=1\linewidth]{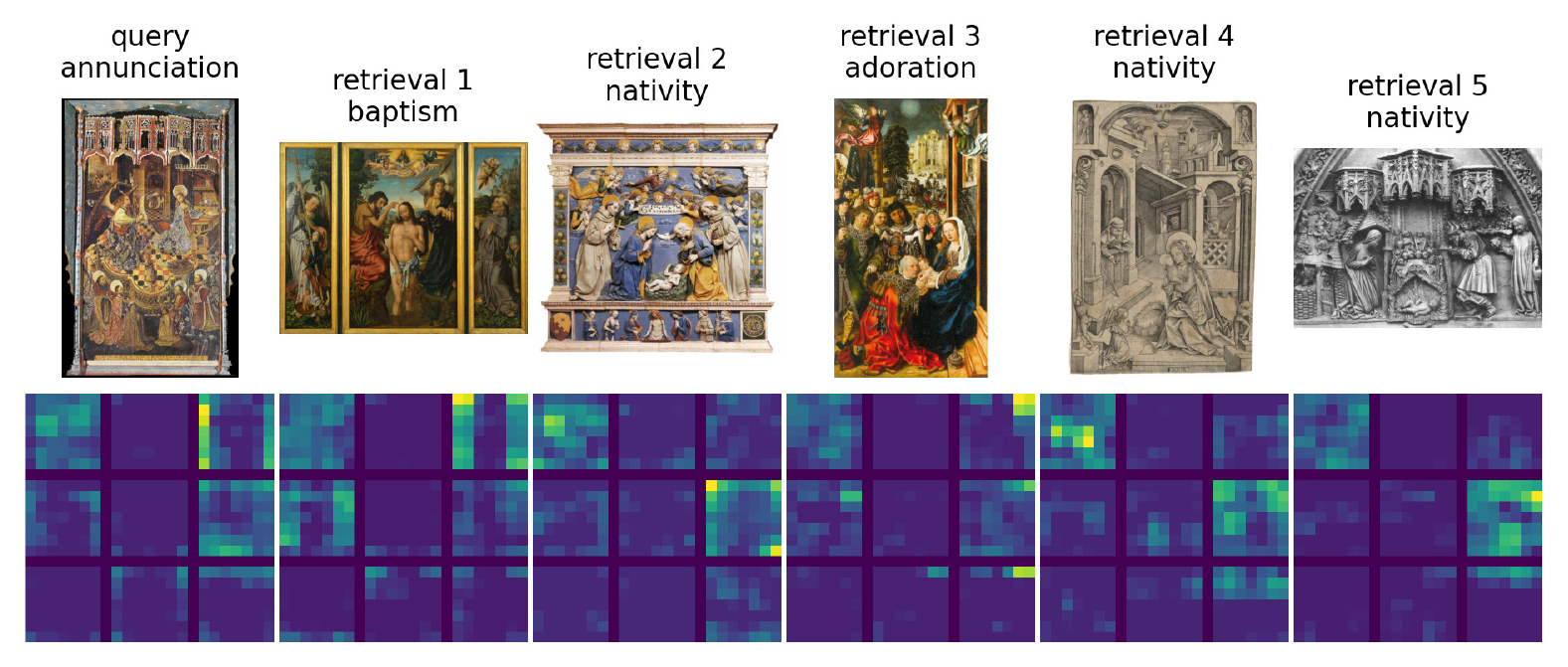}
			\subcaption{VGG19 (worst)}
			\label{fig:eval:plots_vgg19_all_bad}
		\end{subfigure}%
	
		\caption{Top row: Top-5 retrieval results; bottom row: corresponding nine most similar VGG19 feature maps for the respective images above. \subref{fig:eval:plots_vgg19_all} \textbf{VGG19 best results}, \subref{fig:eval:plots_vgg19_all_bad} \textbf{VGG19 worst results} ($\bigoplus$: concat operation).}
	\end{figure}

\textbf{VGG19} retrieval results are depicted in \cref{fig:eval:plots_vgg19_all}. 
We notice that not only the image style plays an important role, but the composition similarity is also captured well. By comparing the query image with the five closest matches, one can quickly notice that the composition with the query is pretty similar, \eg in terms of the two persons and their relation between them while the style is not very dominant. Generally, it is difficult to understand the decision process of very deep networks because of the large feature space. In an attempt to understand the retrieval process, we selected the $9$ (out of total $512$) most similar feature maps from VGG19 (bottom row in \cref{fig:eval:plots_vgg19_all}). The first row and second column seems to have high activations in the regions where the arcs are located which can be considered a good feature for the compositional similarity. For the five worst retrieval results shown in \cref{fig:eval:plots_vgg19_all_bad}, the style similarity seems more dominant for the decision process and the retrieved results are therefore relatively weak.  
	
	\begin{figure}[t]
		\begin{subfigure}{0.95\textwidth}
			\centering
			\includegraphics[width=\linewidth]{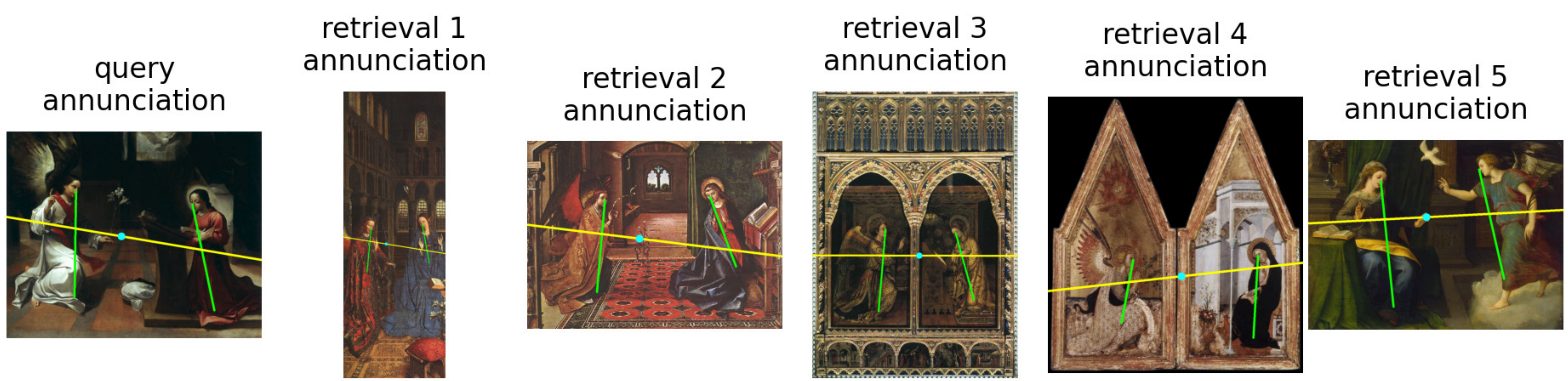}
			\subcaption{\icc U -- best}
			\label{fig:eval:plots_icc_t_ar_best}
		\end{subfigure}
		\\
		\begin{subfigure}{0.95\textwidth}
			\centering
			\includegraphics[width=\linewidth]{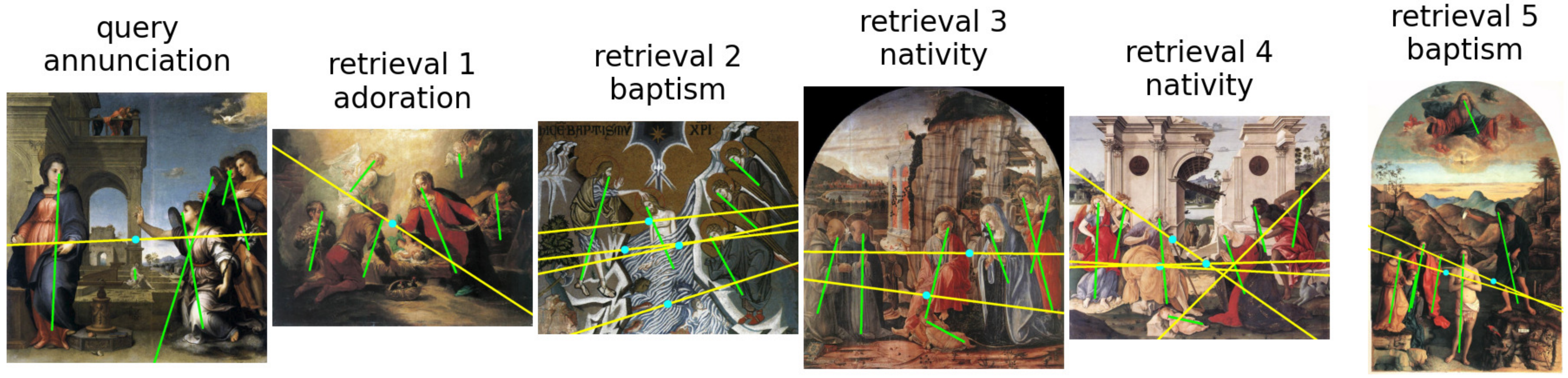}
			\caption{\icc U -- worst}
			\label{fig:eval:plots_icc_t_ar_bad}
		\end{subfigure}%

		\caption{\textbf{\icc U retrieval results} \subref{fig:eval:plots_icc_t_ar_best} best and \subref{fig:eval:plots_icc_t_ar_bad} worst results overlayed with compositional elements. The query image contains all detected poses and the target images only the pose with the smallest distance out of all bipartite combinations between query and retrieved image.}
	\end{figure}

	\begin{figure}[t]
		\centering
		\begin{subfigure}{0.95\textwidth}
			\includegraphics[width=\linewidth]{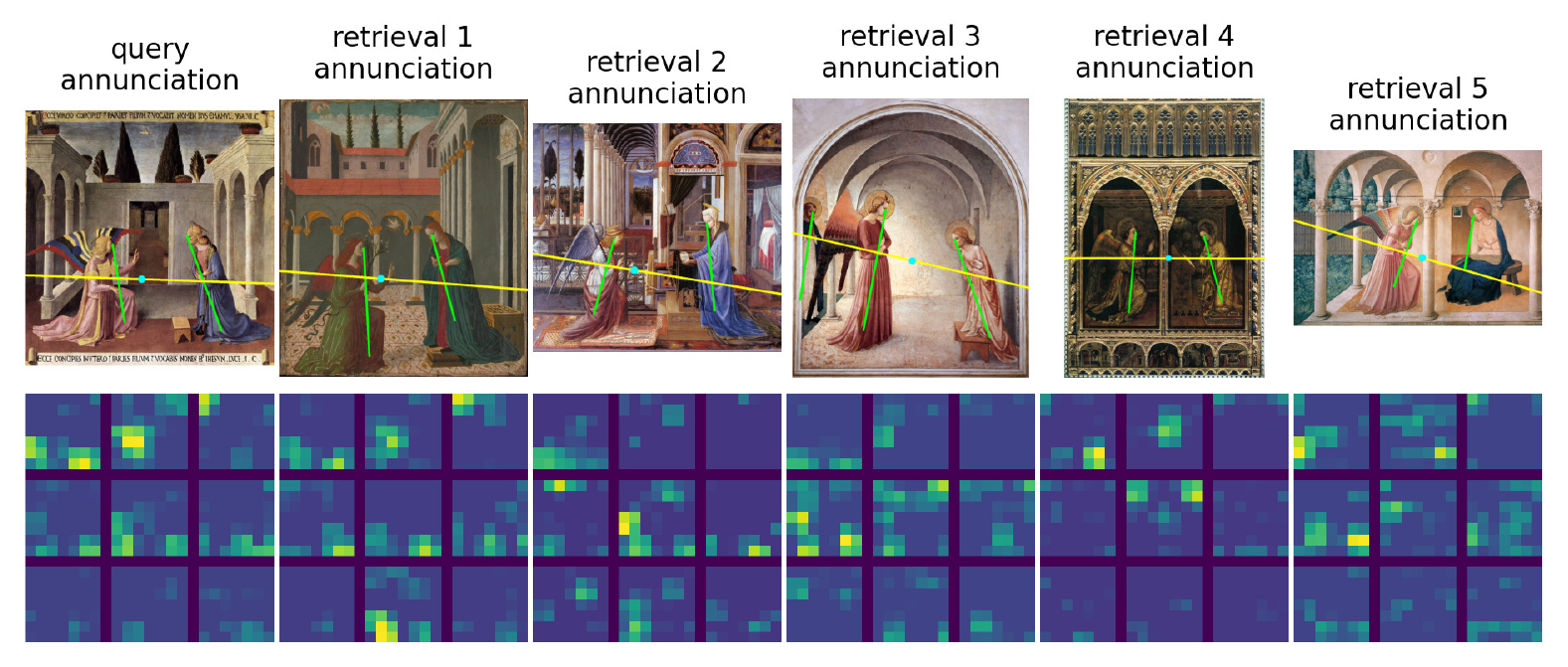}
			\subcaption{\icc U $\bigoplus$ VGG19 (best)}
			\label{fig:eval:plots_icc_combined}
		\end{subfigure}%
        \\
		\begin{subfigure}{0.95\textwidth}
			\centering
			\includegraphics[width=\linewidth]{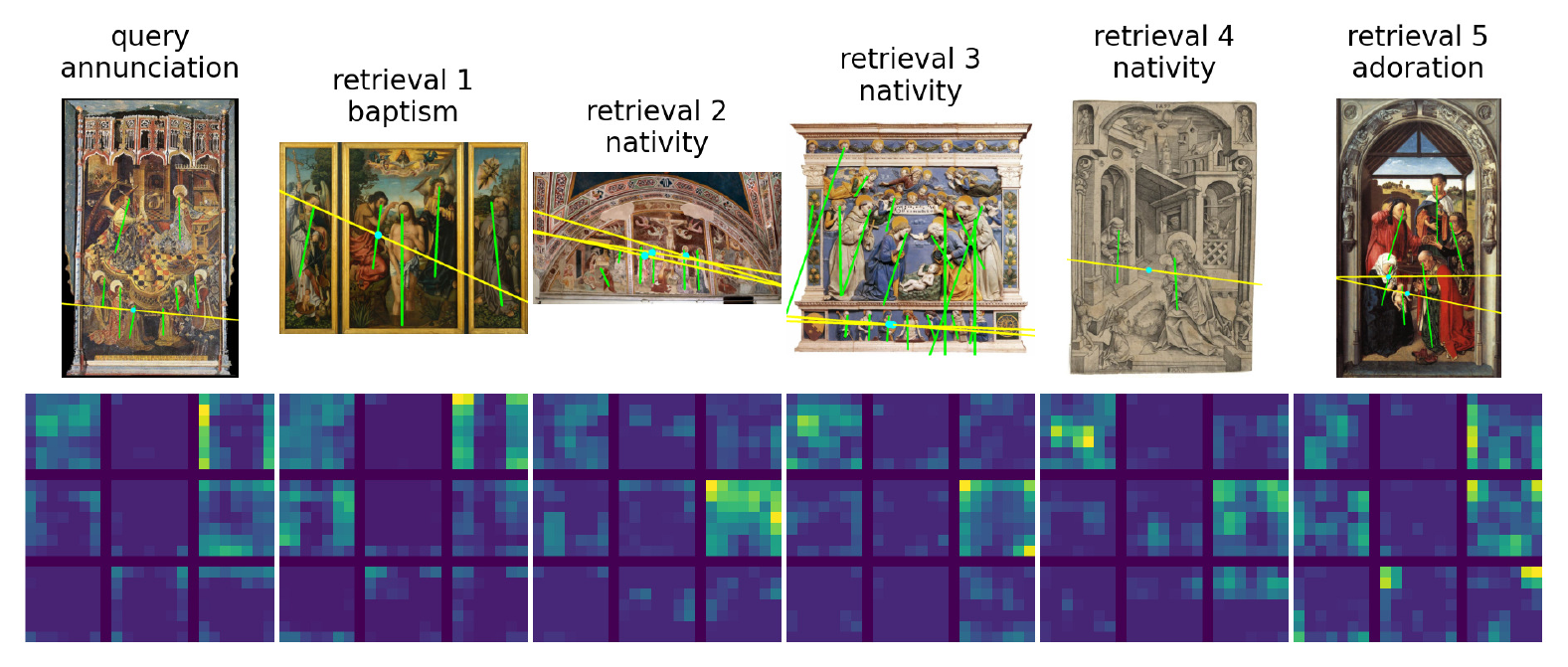}
			\subcaption{\icc U $\bigoplus$ VGG19 (worst)}
			\label{fig:eval:plots_icc_combined_bad}
		\end{subfigure}
	
		\caption{Top row: Top-5 retrieval results; bottom row: corresponding nine most similar VGG19 feature maps for the respective images above. \subref{fig:eval:plots_icc_combined} \textbf{\icc U $\bigoplus$ VGG19} best retrieval results measured by p@10, \subref{fig:eval:plots_icc_combined_bad} \textbf{\icc U $\bigoplus$ VGG19} worst retrieval results measured by p@10 ($\bigoplus$: concat operation).}
	\end{figure}
	
	\begin{figure}[t]
		\begin{subfigure}{0.95\textwidth}
			\centering
			\includegraphics[width=\linewidth]{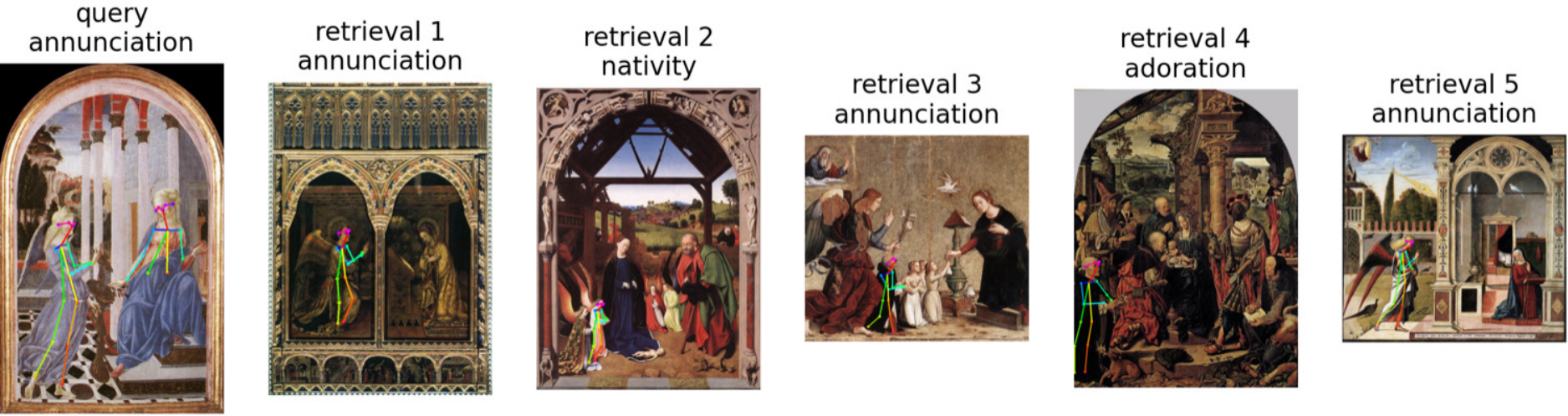}
			\subcaption{LATP -- best}
			\label{fig:eval:plots_latp_best}
		\end{subfigure}
		\\
		\begin{subfigure}{0.95\textwidth}
			\centering
			\includegraphics[width=\linewidth]{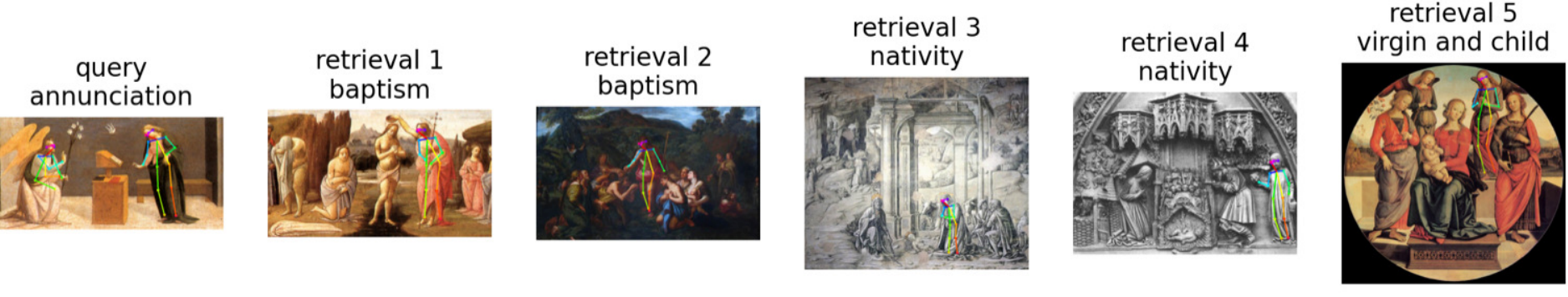}	
			\subcaption{LATP -- worst}
			\label{fig:eval:plots_latp_bad}
		\end{subfigure}
	
		\caption{\textbf{LATP retrieval results} \subref{fig:eval:plots_latp_best} best and \subref{fig:eval:plots_latp_bad} worst results overlayed with pose skeletons. The query image contains all detected poses and the target images only the pose with the smallest distance out of all bipartite combinations between query and retrieved image.}
	\end{figure}

	\textbf{\icc U}: As seen in \cref{fig:eval:plots_icc_t_ar_best}, our untuned \icc features (\icc U) can represent the composition in the images pretty well, and at the same time are very low dimensional and explainable. In the case of very complex compositions, our few compositional elements are not able to cover the composition well enough and therefore the retrieval method retrieves incorrect results. An example of such a complex composition can be seen in \cref{fig:eval:plots_icc_t_ar_bad}.

	\textbf{\icc U $\bigoplus$ VGG19}: When \icc is combined with deep features, the combined features can tackle such complex compositions. This combined feature exhibits explainability (due to \icc) for the compositions and robustness to style (due to deep features). As seen in \cref{fig:eval:plots_icc_combined}, by combining both methods the retrieval result can still easily be understood from an art historian's perspective with the \icc compositional elements and at the same time is now able to detect the composition in a more reliable manner across the styles. On the other hand, the combined method fails if the visual style is too dominating, as can be observed in \cref{fig:eval:plots_icc_combined_bad}.

	In the \textbf{LATP} retrieval results, shown in \cref{fig:eval:plots_latp_best}, the similarity between the pose structure without the context and position information of the pose is insufficient to find images with the same iconography and likely similar compositions. The same can be observed when looking at the worst retrieval result in \cref{fig:eval:plots_latp_bad}.
	
	In addition to the above discussion, we compare \icc with ICC in the Supplementary (Sec. 7.6)
	with reference to the user study done by Madhu \etal~\cite{madhu2020understanding}. \icc shows promising improvements, especially in terms of poselines, and can cope well with the expert annotators and compare well with ICC.
	
	\subsection{Cluster Analysis}
\begin{figure}[t]
    \centering
	\begin{subfigure}{0.49\textwidth}
		\centering
		\includegraphics[width=0.75\linewidth]{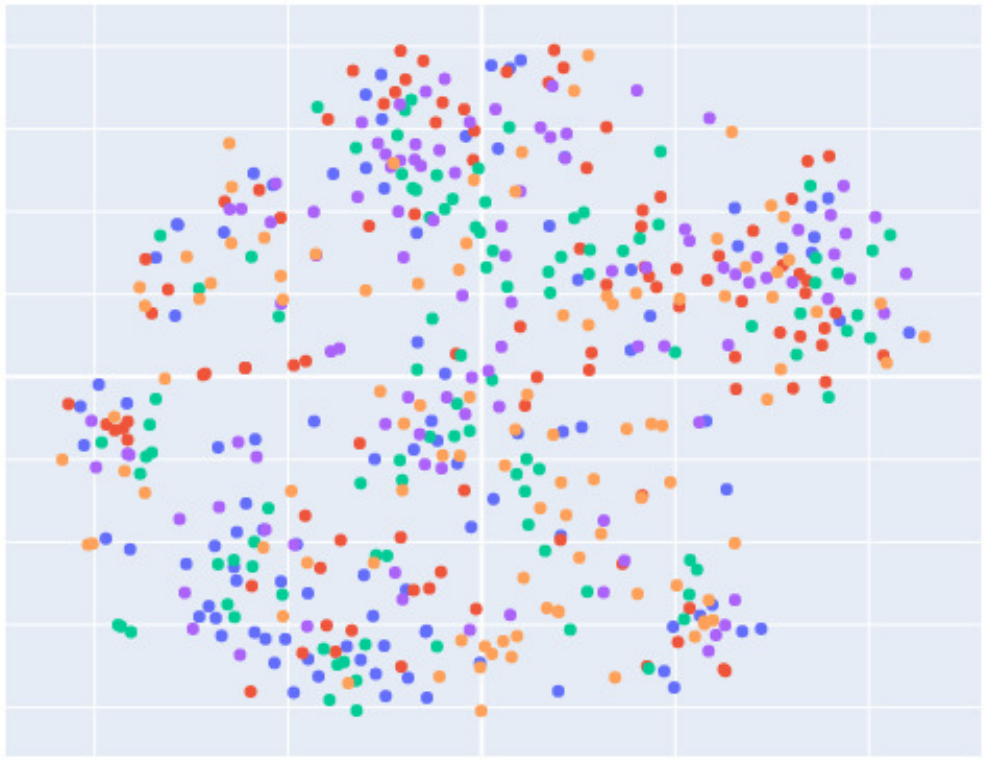}
		\subcaption{$SIFT_{VLAD}$}
		\label{fig:clustering_sift}
	\end{subfigure}
	\centering
	\begin{subfigure}{0.49\textwidth}
		\centering
		\includegraphics[width=0.75\linewidth]{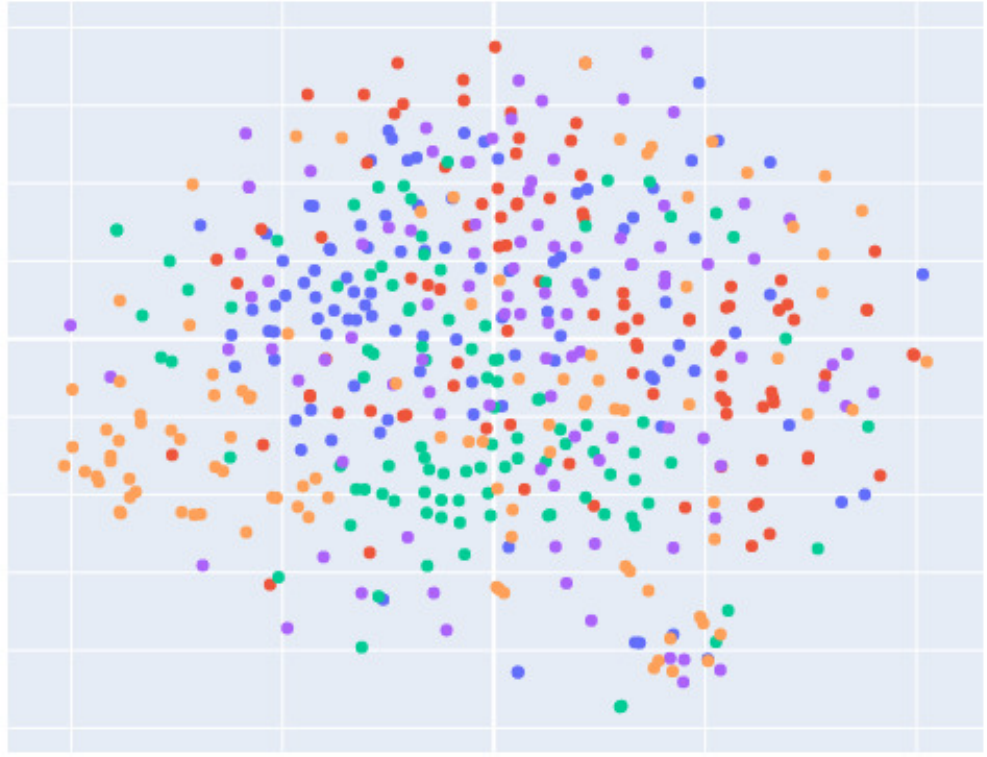}
		\subcaption{VGG19}
		\label{fig:clustering_vgg19}
	\end{subfigure}
	\centering
	\begin{subfigure}{0.49\textwidth}
		\centering
		\includegraphics[width=0.75\linewidth]{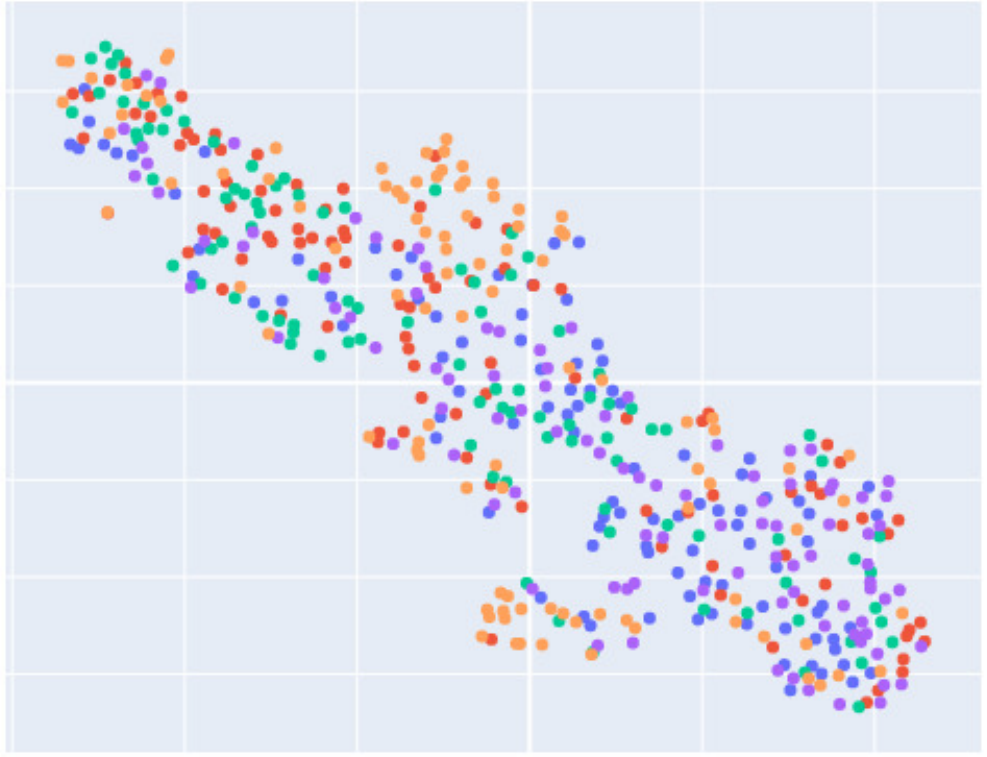}	
		\subcaption{LATP}
		\label{fig:clustering_latp}
	\end{subfigure}
	\centering
	\begin{subfigure}{0.49\textwidth}
		\centering
		\includegraphics[width=0.75\linewidth]{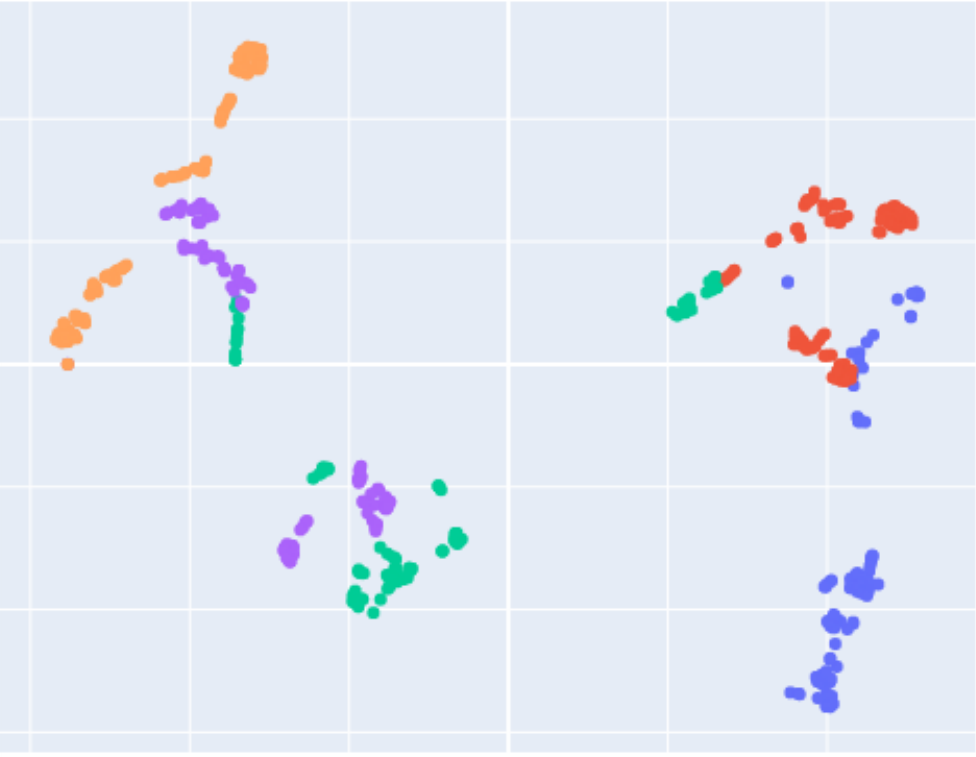}	
		\subcaption{\icc U}
		\label{fig:clustering_icc}
	\end{subfigure}

	\caption{\textbf{t-SNE plots.} Cluster analysis of traditional (SIFT), deep-learning (VGG19), SOTA (LATP), and \icc U methods using first two components; class-color map: adoration (blue), annunciation (red), baptism (green), nativity (purple), virgin and child (orange).}
	\label{fig:clustering}
\end{figure}

The motivation for this work is to generate an interpretable representation of an artwork from a compositional perspective. 
To understand this, we cluster the \icc U features with traditional, deep learning features, and LATP features to visualize the robustness of \icc in the representational space.  
We used the embeddings of the penultimate layer as features of dimension 2048 for ResNet50 and dimension 4096 for VGG19. 
We used the VLAD encodings for SIFT, BRIEF, and ORB features.
Given that $N$ persons and $M$ global action lines are detected, each image is represented by \num[parse-numbers=false]{N x 2 x 2} (action region-normalized poselines) + \num[parse-numbers=false]{M x 2 x 2} (global action lines) keypoints.
All these keypoints are merged into an array, and a 7-D statistical feature vector is generated using mean, standard deviation, median, and percentiles (5, 25, 75, 95).
Similarly, for LATP, where each image is represented by the \num[parse-numbers=false]{N x 18 x 2} keypoints, all these keypoints are merged, and a 7-D statistical feature vector is calculated as described above. 
For clustering and visualization, we generate UMAP and t-SNE plots (\cf \cref{fig:clustering}) for each of the feature representations using the first and the second component.

We can observe in \cref{fig:clustering} that the class instances are strongly clustered for \icc U in comparison to any of the methods, and hence can conclude that the AR-normalized \icc U feature representation is the most robust and explainable feature representation for this dataset. 
Our two-fold reasoning can explain why \icc U features cluster better than others: the action regions in iconography are very similar from a compositional perspective, irrespective of the styles and colors in the painting. 
Using the AR features to normalize local poselines also creates a commonality between the instances of the same iconography. Hence, they seem to cluster better compared to the other features. 

\section{Conclusions and Future Work}\label{sec:conclusion}
The following points summarize the \emph{main conclusions} that can be drawn from our work: 
    \begin{itemize}
    	\item Motivated by Imdahl's ideas on image compositions in art history, we developed a robust and efficient framework to understand compositions in a standardized manner. It is easy to interpret and thus relatively simple to compare with art historical approaches. 
    	\item Our work constitutes a conceptional shift, where the highly criticized compositional analysis tries to find understanding of the artist's intention and the consequent treatment of image space, in the pursuit to finding aesthetic value of the artwork. We can find certain visual features which make the motif comparable, retrievable and addressable for further interpretation. 
    	\item We can also conclude that Image compositions are very useful features for the community since they are light-weight, easily interpretable and visualized. The lines added by \icc describe (inter-)action in the image. They do not (or just by chance) re-imagine lines that were already part of the artistic process. 
    	\item In this manuscript, we present \icc as a tool assisting towards a semantic image understanding for art history improving our previous ICC method in terms of robust poselines abstraction and obtaining correct bisection angles. 
    	\item We focus on the use of the resulting composition diagrams (\icc features) for content-based retrieval of paintings with similar image compositions. Manual generation and matching of \icc features for a large corpus of data is infeasible. Therefore our method has potential use cases in linking big art historical datasets based on their compositions. 
    	\item In addition to its efficiency, our method is also explainable as the retrieved images can be reasoned based on compositional elements which is often missed by the state of the art methods. We showed that our method outperforms other state-of-the-art methods in compositional similarity, while also consistently improving performance when combined with deep features. 
    \end{itemize}
    
    As one weakness of our compositional features is that they are conceptualized as straight lines. Hence, future work can be stated as follows:
    \begin{itemize}
    	\item As described in ~\cref{sec:intro}, the incorporation of more complex compositional elements like contours or connected components could potentially improve the \icc retrieval framework.
    	\item Explainable deep methods are still a work in progress and the compositional features can be integrated with these deep methods at the modeling stage. 
    	\item Training of pose estimation networks on art historical images should also indirectly improve the compositional features.
    	\item The problem with most of the existing SOTA gaze estimation techniques is that they work on video data or they need gaze annotated data. Generating the gaze annotations for art historical corpora and training existing SOTA method can improve the area of focus for the central characters of the image. 
    	\item All the recent techniques focus on end-to-end approach for any task at hand. Designing an end-to-end deep approach for the task of generating image compositions is a highly challenging future goal. 
    \end{itemize}

\newpage
\bibliography{mybibfile}

\newpage

\end{document}